\pdfoutput=1
\documentclass{article}
\PassOptionsToPackage{numbers, compress}{natbib}
\usepackage[final]{nips_2017}
\usepackage{amsmath}
\usepackage{amsfonts}
\usepackage{amssymb}
\usepackage{rbase}
\usepackage{times}

\usepackage{tablefootnote}
\usepackage{booktabs}
\usepackage{hyperref}
\usepackage{graphicx}
\usepackage{subfigure}
\usepackage{xypic}
\usepackage[all,2cell,ps]{xy}
\usepackage{tikz}
\usepackage{epsfig}
\usepackage{enumerate}
\usepackage{stmaryrd}
\usepackage{bbm}
\usepackage{algorithm}
\usepackage{algorithmic}
\usepackage{paralist}
\usepackage{multicol}
\usepackage{soul}
\usepackage{adjustbox}

\newcommand{\diag}{\mathop{\textrm{diag}}}

\newtheorem{prop}{Proposition}

\newtheorem{defn}{Definition}

\newenvironment{pf}{\textbf{Proof.}}{\mbox{}\hfill\m{\blacksquare}\\} 

\newcounter{exmpl}

\newenvironment{mequation*}{$}{$}

\newcommand{\bl}[1]{\llbracket #1 \rrbracket}

\renewcommand{\df}[1]{\textbf{#1}}

\makeatletter
\providecommand*{\cupdot}{\mathbin{\mathpalette\@cupdot{}}}
\newcommand*{\@cupdot}[2]{\ooalign{$\m@th#1\cup$\cr\hidewidth$\m@th#1\cdot$\hidewidth}}
\makeatother

\makeatletter
\def\thm@space@setup{%
  \thm@preskip=-3pt \thm@postskip=5pt
}
\makeatother

\title{Multiresolution Kernel Approximation for \\ Gaussian Process Regression}

\author{Yi Ding\m{{}^\ast}, Risi Kondor\m{{}^{\ast\dag}}, Jonathan Eskreis-Winkler\m{{}^\dag} \\ 
\m{{}^\ast}Department of Computer Science, \m{{}^\dag}Department of Statistics \\
The University of Chicago, Chicago, IL, 60637 \\
\texttt{\{dingy,risi,eskreiswinkler\}@uchicago.edu}}

\begin{document}
\maketitle
\vspace{-10pt}

\newcommand{\tikmxA}[1]{
\br{\,\begin{tikzpicture}[baseline=-15.5, scale=0.055]
\filldraw[gray] (0,0) rectangle +(#1,-#1); \end{tikzpicture}\,}}

\newcommand{\tikmxB}[2]{
\br{\,\begin{tikzpicture}[baseline=-15.5, scale=0.055] 
\filldraw[gray] (0,0) rectangle +(#1,-#1);
\foreach \i in {#1,...,#2}{
\filldraw[gray] (\i,-\i) rectangle +(1,-1);}
\end{tikzpicture}\,}}

\newcommand{\tikmxC}[2]{
\br{\,\begin{tikzpicture}[baseline=-15.5, scale=0.055]
\draw (0,0) rectangle +(17,-17);\draw (10,-9) node {#1}; \end{tikzpicture} \,}}

\newcommand{\tikmxAb}[1]{
\br{\,\begin{tikzpicture}[baseline=-12.5, scale=0.07]
\filldraw[gray] (0,0) rectangle +(#1,-#1); \end{tikzpicture}\,}}

\newcommand{\tikmxBb}[2]{
\br{\,\begin{tikzpicture}[baseline=-13, scale=0.07] 
\filldraw[gray] (0,0) rectangle +(#1,-#1);
\foreach \i in {#1,...,#2}{
\filldraw[gray] (\i,-\i) rectangle +(1,-1);}
\end{tikzpicture}\,}}

\begin{abstract}
Gaussian process regression generally does not scale to beyond a few thousands data points
without applying some sort of kernel approximation method. 
Most approximations focus on the high eigenvalue part of the spectrum of the kernel matrix, $K$, 
which leads to bad performance when the length scale of the kernel is small. 
In this paper we introduce Multiresolution Kernel Approximation (MKA), the first true broad bandwidth 
kernel approximation algorithm. 
Important points about MKA are that it is memory efficient, and it is a direct method, 
which means that it also makes it easy to approximate $K^{-1}$ and $\mathop{\textrm{det}}(K)$. 
\end{abstract} 
\section{Introduction}


Gaussian Process (GP) regression, and its frequentist cousin, kernel ridge regression,  
are such natural and canonical algorithms that they have been reinvented many 
times by different communities under different names. 
In machine learning, GPs are considered one of the standard methods of Bayesian nonparametric inference 
\cite{RasmussenBook}. 
Meanwhile, the same model, under the name Kriging or Gaussian Random Fields, is the de facto standard for  
modeling a range of natural phenomena from geophyics to biology   
\cite{stein1999}. 
One of the most appealing features of GPs is that, ultimately, 
the algorithm reduces to ``just'' having to compute the inverse of a kernel matrix, \m{K}.  
Unfortunately, this also turns out to be the algorithm's Achilles heel, since 
in the general case, the complexity of inverting a dense \m{n\<\times n} matrix 
scales with \m{O(n^3)}, 
meaning that when the number of training examples exceeds 
\m{10^4\<\sim 10^5}, GP inference becomes problematic on virtually any computer\footnote{
In the limited case of evaluating 
a GP with a fixed Gram matrix on a single training set, GP inference reduces 
to solving a linear system in \m{K}, which scales better with \m{n}, but 
might be problematic behavior when the condition number of \m{K} is large.}. 
Over the course of the last 15 years, devising approximations to address this problem 
has become a burgeoning field. 

The most common approach 
is to use one of the so-called Nystr\"om methods \cite{Williams2001}, 
which select a small subset \m{\cbr{x_{i_1},\ldots,x_{i_m}}} of the original training data points as ``anchors''  
and approximate \m{K} in the form 
\m{K\approx K_{\ast,I} C K_{\ast,I}^\top},
where \m{K_{\ast,I}} is the submatrix of \m{K} consisting of columns \m{\cbr{\sseq{i}{m}}}, and \m{C} is a 
matrix such as the pseudo-inverse of \m{K_{I,I}}. 
Nystr\"om methods often work well in practice and have a mature literature offering strong theoretical guarantees. 
Still, Nystr\"om is inherently a global low rank approximation, and, as pointed out in \cite{Si2014b}, 
a priori there is no reason to believe that \m{K} should be well approximable by a low rank matrix: 
for example, in the case of the popular Gaussian kernel \m{k(x,x')=\exp(-(x\<-x')^2/(2\ell^2))}, 
as \m{\ell} decreases and the kernel becomes more and more ``local'' the number of significant 
eigenvalues quickly increases. 
This observation has motivated alternative types of approximations, including local, hierarchical 
and distributed ones (see Section \ref{sec: approx}). 
In certain contexts involving translation invariant kernels 
yet other strategies may be applicable \cite{Rahimi2008}, but these are 
beyond the scope of the present paper. 

In this paper we present a new kernel approximation method, Multiresolution Kernel Approximation (MKA), 
which is inspired by a combination of ideas from hierarchical 
matrix decomposition algorithms and multiresolution analysis. 
Some of the important features of MKA are that (a) it is a broad spectrum algorithm that 
approximates the entire kernel matrix \m{K}, not just its top eigenvectors, and 
(b) it is a so-called ``direct'' method, i.e., it yields explicit  
approximations to \m{K^{-1}} and \m{\mathop{\mathrm{det}}(K)}. 
\vspace{-8pt}




\paragraph{Notations.}
We define \m{[n]\<=\cbrN{\oneton{n}}}. 
Given a matrix \m{A}, 
and a tuple \m{I\<=\br{\sseq{i}{r}}}, \m{A_{I,\ast}} will denote the submatrix of \m{A} formed of 
rows indexed by \m{\sseq{i}{r}}, similarly \m{A_{\ts\ast,J}} will denote the submatrix formed of columns indexed 
by \m{\sseq{j}{p}}, and \m{A_{I,J}} will denote the submatrix at the intersection of rows 
\m{\sseq{i}{r}} and columns \m{\sseq{j}{p}}. 
We extend these notations to the case when \m{I} and \m{J} are sets in the obvious way. 
If \m{A} is a blocked matrix then \m{\bl{A}_{i,j}} will denote its \m{(i,j)} block.  

\section{Local vs. global kernel approximation}\label{sec: approx}

Recall that a Gaussian Process (GP) on a space \m{\Xcal} is a prior over 
functions \m{f\colon \Xcal\to\RR} defined by a mean function \m{\mu(x)=\mathbb{E}[f(x)]},  
and covariance function 
\m{k(x,x')=\Cov(f(x),f(x'))}. 
Using the most elementary model    
\m{y_i = f(x_i)+\epsilon} 
where \m{\epsilon\sim \Ncal(0,\sigma^2)} and \m{\sigma^2} is a noise parameter, 
given training data \m{\cbr{(x_1,y_1),\ldots,(x_n,y_n)}}, 
the posterior is also a GP, with mean 
\begin{mequation} 
\mu'(x)=\mu(x)+\V k^{\top}_x (K+\sigma^2 I)^{-1} \V y,
\end{mequation}
where \m{\V k_x\<=(k(x,x_1),\ldots,k(x,x_n)),\,} \m{\V y\!=\!(\sseq{y}{n})}, and covariance 
\begin{equation}\label{eq: gp pvar}
k'(x,x')=k(x,x')-\V k^{\top}_{x'} (K+\sigma^2 I)^{-1} \V k_{x}. 
\end{equation}
Thus (here and in the following assuming \m{\mu\<=0} for simplicity), 
the maximum a posteriori (MAP) estimate of \m{f} is \vspace{-5pt}
\begin{equation}\label{eq: gp map}
\h f(x)=\V k^{\top}_x (K+\sigma^2 I)^{-1} \V y. 
\end{equation}
Ridge regression, which is the frequentist analog of GP regression, 
yields the same formula, but regards 
\m{\h f} as the solution to a 
regularized risk minimization problem over a Hilbert space \m{\Hcal} induced by \m{k}. 
We will use ``GP'' as the generic term to refer to both Bayesian GPs 
and ridge regression.
\ignore{
Letting \m{K'=(K\<+\sigma^2 I)}, the computational challenge for both 
algorithms is evaluating \m{K'^{-1} \V y} in \rf{eq: gp map} (or \m{K'^{-1} \V k_x} in \rf{eq: gp pvar}) as 
the number of training examples, \m{n}, gets large. 
When the GP is to be run only once, with a fixed \m{\V y} vector, 
often the best approach is to solve the linear system \m{K' \V\alpha\<=\V y} 
with an iterative solver, such as conjugate gradients. 
In most cases this is much faster than computing \m{{K'}^{-1}} explicity, however care must be taken 
because the convergence rate of such methods is typically inversely proportional to the condition 
number of \m{K'} (granted, the \m{\sigma^2 I} ``regularization term'' does help somewhat in this regard). 
In many cases, however, a given GP needs to be solved for many different \m{\V y} vectors, or, e.g., 
in multidimensional regression or multiclass classification, each \m{y_i} might itself be 
multidimensional and then, computing the \m{K'^{-1}} explicitly is sometimes not unreasonable. 
}
Letting \m{K'=(K\<+\sigma^2 I)}, virtually all GP approximation approaches  
focus on trying to 
approximate the (augmented) kernel matrix \m{K'} in such a way so as to make inverting it, solving 
\m{K'\V y\<=\V\alpha} or computing \m{\det(K')} easier. 
For the sake of simplicity in the following we will actually discuss approximating \m{K}, 
since adding the diagonal term usually doesn't make the problem 
any more challenging. 

\subsection{Global low rank methods}

As in other kernel methods, intuitively, \m{K_{i,j}\<=k(x_i,x_j)} encodes the degree of similarity or 
closeness between the two points \m{x_i} and \m{x_j} as it relates to the degree of correlation/similarity  
between the value of \m{f} at \m{x_i} and at \m{x_j}. 
Given that \m{k} is often conceived of as a smooth, slowly varying function, 
one very natural idea is to take a smaller set 
\m{\cbr{x_{i_1},\ldots,x_{i_m}}} of ``landmark points'' or ``pseudo-inputs''
and approximate \m{k(x,x')} in terms of the similarity of \m{x} to each of the landmarks, 
the relationship of the landmarks to each other, and the similarity of the landmarks to \m{x'}. 
Mathematically,  
\[k(x,x')\approx \sum_{s=1}^{m} \sum_{j=1}^{m} k(x,x_{i_s})\,c_{i_s,i_j}\,k(x_{i_j},x'),\]
which, assuming that \m{\cbr{x_{i_1},\ldots,x_{i_m}}} is a subset of the original point set 
\m{\cbr{\sseq{x}{n}}}, amounts to an approximation of the form 
\m{K\approx K_{\ast,I} \ts C\ts K_{\ast,I}^\top},
with \m{I=\cbr{\sseq{i}{m}}}. 
The canonical choice for \m{C} is \m{C\<=W^+}, where 
\m{W\<=K_{I,I}},
and \m{{W}^{+}} denotes the Moore-Penrose pseudoinverse of \m{W}. 
The resulting approximation 
\begin{equation}\label{eq: Nystrom2}
K\approx K_{\ast,I} W^{+} K_{\ast,I}^\top,
\end{equation}
is known as the Nystr\"om approximation, because it is analogous to the so-called Nystr\"om extension 
used to extrapolate continuous operators from a finite number of quadrature points. 
Clearly, the choice of \m{I} is critical for a 
good quality approximation. 
Starting with the pioneering papers \cite{SmolaGreedy,Williams2001,Fowlkes2004}, over the course 
of the last 15 years a sequence of different sampling strategies have been developed for 
obtaining \m{I}, 
several with rigorous approximation bounds \cite{Drineas2005,RongJin2013,Gittens2013,Sun2015}. 
Further variations include the ensemble Nystr\"om method 
\cite{Kumar2009} and the modified Nystr\"om method \cite{Wang2014}. 

Nystr\"om methods have the advantage of being relatively simple, and having reliable performance bounds. 
A fundamental limitation\ignore{of the Nystr\"om approach}, however, is that the approximation 
\rf{eq: Nystrom2} is inherently low rank. 
As pointed out in \cite{Si2014b}, 
there is no reason to believe that kernel matrices in general should be close to low rank.
An even more fundamental issue, which is less often discussed in the literature, relates to the specific form of 
\rf{eq: gp map}. The appearance of \m{K'^{-1}} in this formula suggests that it is the \emph{low} eigenvalue eigenvectors 
of \m{K'} that should dominate the result of GP regression. 
On the other hand, multiplying the matrix by \m{\V k_x} largely cancels this effect, since \m{\V k_x} is 
effectively a row of a kernel matrix similar to \m{K'}, and will likely concentrate most weight on the \emph{high} 
eigenvalue eigenvectors. Therefore, ultimately, it is not \m{K'} itself, but the relationship 
between the eigenvectors of \m{K'} and the data vector \m{\V y} that determines which part of 
the spectrum of \m{K'} the result of GP regression is most sensitive to. 

Once again, intuition about the kernel helps clarify this point. In a setting where the function that 
we are regressing is smooth, and correspondingly, the kernel has a large length scale parameter, 
it is the global, long range relationships between data points 
that dominate GP regression, and that can indeed be well approximated by the landmark point method. 
In terms of the linear algebra, the spectral expansion of \m{K'} is dominated by a 
few large eigenvalue eigenvectors, 
we will call this the ``PCA-like'' scenario.
In contrast, in situations where \m{f} varies more rapidly, a shorter lengthscale kernel is 
called for, local relationships between nearby points become more important, which the landmark point method 
is less well suited to capture.  
We call this the ``\m{k}--nearest neighbor type'' scenario. 
In reality, most non-trivial GP regression problems fall somewhere in between the above two extremes.  
In high dimensions data points tend to be all almost equally far from each other anyway, 
limiting the applicability of simple geometric interpretations. 
Nonetheless, the two scenarios are an illustration of the general point that 
one of the key challenges in large scale machine learning is integrating  
information from both local and global scales.  
\vspace{-5pt}

\subsection{Local and hierarchical low rank methods}\label{sec: local}

Realizing the limitations of the low rank approach, local kernel approximation methods have also started 
appearing in the literature. 
Broadly, these algorithms: 
(1) first cluster the rows/columns of \m{K} with some appropriate fast clustering method, 
e.g., METIS \cite{metis} or GRACLUS \cite{Dhillon2007} and block \m{K} accordingly; 
(2) compute a low rank, but relatively high accuracy, approximation \m{\bl{K}_{i,i}\approx U_i \Sigma_i U_i^\top} to 
each diagonal block of \m{K}; 
(3) use the \m{\cbrN{U_i}} bases to compute possibly coarser approximations to the 
\m{\bl{K}_{i,j}} off diagonal blocks. 
This idea appears in its purest form in \cite{Savas}, and is refined in \cite{Si2014b} 
in a way that avoids having to form all rows/columns of the off-diagonal blocks in the first place. 
Recently, \cite{Wang2015} proposed a related approach, where all the blocks in a given row share the same 
row basis but have different column bases. 
A major advantage of local approaches is that they are inherently parallelizable. 
The clustering itself, however, is a delicate, and sometimes not very robust component of these methods. 
In fact, divide-and-conquer type algorithms such as \cite{Balcan2014} and \cite{Zhang2013} can also 
be included in the same category, even though in these cases the blocking is usually random. 


A natural extension of the blocking idea would be to apply the divide-and-conquer approach recursively, at 
multiple different scales. 
Geometrically, this is similar to recent multiresolution data analysis  
approaches such as \cite{allard2012multi}. 
In fact, hierarchical matrix approximations, including HODLR matrices, 
\m{\mathcal{H}}--matrices \cite{Hackbusch1999}, 
\m{\mathcal{H}^2}--matrices \cite{Hackbusch1999b} and HSS matrices \cite{Chandrasekaran2005} 
are very popular in the numerical analysis literature.
While the exact details vary, each of these methods imposes a specific type of block structure 
on the matrix and forces the off-diagonal blocks to be low rank   
(Figure 1 in the Supplement). 
Intuitively, nearby clusters interact in a richer way, but as 
we move farther away, data can be aggregated more and more coarsely, just as in the fast multipole method 
\cite{Greengard1987}.

We know of only two applications of the hierarchical matrix methodology to kernel approximation: 
B\"orm and Garcke's \m{\Hcal^2} matrix approach \cite{Borm2007} and 
O'Neil et al.'s HODLR method \cite{Ambikasaran2015}. 
The advantage of \m{\Hcal^2} matrices is their more intricate structure, allowing relatively tight 
interactions between neighboring clusters 
even when the two clusters are not siblings in the tree 
(e.g. blocks 8 and 9 in Figure 1c in the Supplement). 
However, the \m{\Hcal^2} format does not directly help with 
inverting \m{K} or computing its determinant: it is merely a memory-efficient way of storing \m{K} 
and performing matrix/vector multiplies \emph{inside} an iterative method. 
HODLR matrices have a simpler structure, but admit a factorization that makes it possible to 
directly compute both the inverse and the determinant of the approximated matrix in just \m{O(n\log n)} time. 

The reason that hierarchical matrix approximations have not become more popular in machine learning so far  
is that in the case of high dimensional, unstructured data, finding the way to organize \m{\cbrN{\sseq{x}{n}}} 
into a single hierarchy is much more challenging than in the setting of regularly spaced points in \m{\RR^2} 
or \m{\RR^3}, where these methods originate:
\begin{inparaenum}[~1.]
\item Hierarchical matrices require making hard assignments of data points to clusters, since the block structure at 
each level corresponds to partitioning the rows/columns of the original matrix. 
\item The hierarchy must form a single tree, which puts deep divisions between clusters 
whose closest common ancestor is high up in the tree. 
\item Finding the hierarchy in the first place is by no means trivial. Most works use 
a top-down strategy which defeats the inherent parallelism of the matrix structure, and the actual 
algorithm used (kd-trees) is known to be problematic in high dimensions \cite{rajani2015}.  
\end{inparaenum}

\section{Multiresolution Kernel Approximation}\label{sec: mmf}
Our goal in this paper is to develop a data adapted multiscale kernel matrix approximation method, 
\df{Multiresolution Kernel Approximation (MKA)}, 
that reflects the ``distant clusters only interact in a low rank fashion'' insight of the 
fast multipole method, but is considerably more flexible than existing hierarchical matrix decompositions. 
The basic building blocks of MKA are local factorizations of a specific form,  
which we call core-diagonal compression. 

\begin{defn}
We say that a matrix \m{H} is \m{\mathbf{c}}\df{--core-diagonal} if \m{H_{i,j}\<=0} unless 
either\, \m{i,j\<\leq c}\, or \m{i\<=j}. 
\end{defn}

\begin{defn}
A \m{\mathbf{c}}\df{--core-diagonal compression} of a symmetric matrix \m{A\tin\RR^{m\times m}} 
is an approximation of the form \vspace{-5pt}
\begin{equation}\label{eq: compr}
A\approx Q^\top H\, Q=
\tikmxAb{10} 
\tikmxBb{5}{10}
\tikmxAb{10},
\end{equation}
where \m{Q} is orthogonal and \m{H} is \m{c}--core-diagonal. 
\end{defn}

Core-diagonal compression is to be contrasted with 
rank \m{c} sketching, where \m{H} would just have the \m{c\<\times c} block, without the rest of the diagonal. 
From our multiresolution inspired point of view, however, 
the purpose of \rf{eq: compr} is not just to sketch \m{A},   
but to also to split \m{\RR^m} into the direct sum of two subspaces: 
(a) the ``detail space'', spanned by the last \m{n\<-c} rows of \m{Q},  
responsible for capturing purely local interactions in \m{A} and  
(b) the ``scaling space'', spanned by the first \m{c} rows,  
capturing the overall structure of \m{A} and its relationship to other diagonal blocks.  

Hierarchical matrix methods apply low rank decompositions to many blocks of \m{K} in parallel, at different scales. 
MKA works similarly, by applying core-diagonal compressions. 
Specifically, the algorithm 
proceeds by taking \m{K} through a sequence of transformations 
\m{K\<=K_0\mapsto K_1 \mapsto \ldots \mapsto K_s}, called stages. 
In the first stage 
\begin{compactenum}[~1.]
\item Similar to other local methods, 
MKA first uses a fast clustering method to cluster the rows/columns of \m{K_0} into clusters  \m{\sseq{\Ccal^1}{p_1}}. 
Using the corresponding permutation matrix \m{C_1} (which maps the elements of the first cluster to 
\m{(1,2,\ldots \absN{\Ccal^1_1})}, the elements of the second cluster to 
\m{(\absN{\Ccal^1_1}\<+1,\ldots,\absN{\Ccal^1_1}\<+\absN{\Ccal^1_2})}, and so on) 
we form a blocked matrix \vspace{-2pt} 
\m{\wbar{K_0}=C_1\ts K_0\ts C_1^\top}, 
where \m{\bl{\wbar{K_0}}_{i,j}=K_{\Ccal^1_i,\Ccal^1_j}}. 
\item Each diagonal block of \m{\wbar{K_0}} is independently core-diagonally 
compressed as in \rf{eq: compr} to yield \vspace{-5pt} 
\begin{equation}\label{eq: cd approx}
H^1_i =\br{Q^1_i\:\bl{\wbar{K_0}}_{i,i}\,(Q^1_i)^\top}_{\textrm{CD}(c^1_i)} 
\end{equation} \vspace{-12pt}\mbox{}\\ 
where \m{CD(c^1_i)} in the index stands for truncation to \m{c^1_i}--core-diagonal form. 
\item The \m{Q^1_i} local rotations are assembled into a single large orthogonal matrix 
\m{\wbar{Q_1}=\bigoplus_{i} Q^1_i} and applied to the full matrix to give 
\m{\wbar{H_1}=\wbar{Q_1}\,\wbar{K_0}\,\wbar{Q_1}^\top}. 
\item The rows/columns of \m{\wbar{H_1}} are rearranged by applying a permutation \m{P_1} that  
maps the core part of each block to one of the first \m{c_1:=c^1_1+\ldots c^1_{p_1}} coordinates, and the 
diagonal part to the rest, giving 
\m{H_1^{\text{pre}}=P_1\:\wbar{H_1}\:P_1^\top}. 
\item Finally, \m{H_1^{\text{pre}}} is truncated into the core-diagonal form 
\m{H_1 = K_1 \oplus D_1}, 
where \m{K_1\tin\RR^{c_1\times c_1}} is dense, while \m{D_1} is diagonal. 
Effectively, \m{K_1} is a compressed version of \m{K_0}, while \m{D_1} is formed  
by concatenating the diagonal parts of each of the \m{H^1_i} matrices. 
Together, this gives a global core-diagonal compression \vspace{-5pt} 
\[K_0\approx \underbrace{C_1^\top  \wbar{Q_1}{}^\top P_1^\top}_{\Qcal_1^\top} 
(K_1\<\oplus D_1)\ts 
\underbrace{P_1\ts \wbar{Q_1}\ts C_1}_{\Qcal_1} \]
\vspace{-15pt}\mbox{}\\ 
of the entire original matrix \m{K_0}. 
\end{compactenum}
The second and further stages of MKA consist of applying the above five steps to \m{K_1,K_2,\ldots, K_{s-1}} in turn,   
so ultimately the algorithm yields a kernel approximation \m{\tilde K} which has a telescoping form 
\begin{equation}\label{eq: mka}
\tilde K\approx 
\Qcal_1^\top\nts (
\Qcal_2^\top\nts (
\ldots 
\Qcal_s^\top\nts (
K_s\<\oplus D_s) \Qcal_s
\ldots 
\<\oplus D_2) \Qcal_2
\<\oplus D_1) \Qcal_1
\end{equation}
The pseudocode of the full algorithm is in the Supplementary Material. 

MKA is really a meta-algorithm, in the sense that it can be used in conjunction with different core-diagonal compressors. The main requirements on the compressor are that 
(a) the core of \m{H} should capture the dominant part of \m{A}, in particular the subspace that most strongly interacts with other blocks,  
(b) the first \m{c} rows of \m{Q} should be as sparse as possible. 
We consider two alternatives. 

\textbf{Augmented Sparse PCA (SPCA).}~ 
Sparse PCA algorithms explicitly set out to find a set of 
vectors \m{\cbrN{\sseq{\V v}{c}}} so as to maximize \m{\nmN{V^\top\!\! A V}_{\text{Frob}}}, 
where \m{V=[\sseq{\V v}{c}]}, 
while constraining each vector to be as sparse as possible \cite{Zou2004}. 
While not all SPCAs guarantee orthogonality, this can be enforced a posteriori via e.g., QR factorization, 
yielding \m{Q_{\text{sc}}}, the top \m{c} rows of \m{Q} in \rf{eq: compr}.  
Letting \m{U} be a basis for the complementary subspace, the optimal choice for 
the bottom \m{m\<-c} rows in terms of minimizing Frobenius norm error of the 
compression is \m{Q_{\text{wlet}}\<=U \hat O}, where 
\[\hat O=\argmax_{O^\top\! O=I} \nmN{\diag(O^\top\!  U^\top\!\! A\, U O)},\] \vspace{-5pt}\\ 
the solution to which is of course given by the eigenvectors of \m{U^\top\!\! A U}. 
The main drawback of the SPCA approach is its computational cost: depending on the 
algorithm, the complexity of SPCA scales with \m{m^3} or worse \cite{Berthet2013,Kuleshov2013}.

\textbf{Multiresolution Matrix Factorization (MMF)}~ 
MMF is a recently introduced matrix factorization algorithm motivated by 
similar multiresolution ideas as the present work, 
but applied at the level of 
individual matrix entries rather than at the level of matrix blocks \cite{MMFicml2014}. 
Specifically, MMF yields a factorization of the form \vspace{-5pt}
\begin{equation*}\label{eq: mmf}
A\approx \underbrace{q_1^\top \ldots q_L^\top}_{Q^\top} H\, \underbrace{q_L\ldots q_1}_Q, 
\end{equation*}
where, in the simplest case, the \m{q_i}'s are 
just Givens rotations.  
Typically, the number of rotations in MMF is \m{O(m)}. 
MMF is efficient to compute, and sparsity is guaranteed by the sparsity of the individual 
\m{q_i}'s and the structure of the algorithm. 
Hence, MMF has complementary strengths to SPCA: 
it comes with strong bounds on sparsity and computation time, but the quality of 
the scaling/wavelet space split that it produces is less well controlled. 


\textbf{Remarks.} We make a few remarks about MKA. 
\begin{inparaenum} 
\ignore{
\item So far, we have left open the issue of how to compute the \rf{eq: compr} local core-diagonal compressions. 
In principle, \rf{eq: compr} could be any low rank approximation of \m{A} augmented with diagonal entries. 
However, this would result in \m{Q} matrices that are dense. In practice we use 
local MMF factorizations of the form defined in \cite{MMFicml2014} because these have the 
dual advantages of (a) being fast to derive (b) leading to \m{Q} matrices that, while dense, are factored 
into a product of \m{O(m)} very sparse rotations. 
}
\item Typically, low rank approximations 
reduce dimensionality quite aggressively. 
In contrast, in core-diagonal compression \m{c} is often on the order of \m{m/2}, leading to ``gentler''  
and more faithful, kernel approximations. 
\item In hierarchical matrix methods, the block structure of the matrix is defined by a single tree, which, 
as discussed 
above, is potentially problematic. In contrast, by virtue 
of reclustering the rows/columns of \m{K_\ell} before every stage, MKA affords a more flexible factorization. 
In fact, beyond the first stage, it is not even individual data points that MKA clusters, but subspaces defined 
by the earlier local compressions. 
\item While 
\m{C_\ell} and \m{P_\ell} are presented as 
explicit permutations, 
they really just correspond to different ways of blocking \m{K_s}, 
which 
is done implicitly in practice with relatively little overhead. 
\item Step 3 of the algorithm is critical, because it extends the core-diagonal splits found in the 
diagonal blocks of the matrix to the off-diagonal blocks. Essentially the same is done in 
\cite{Si2014b} and \cite{Wang2015}. 
This operation reflects a structural assumption about \m{K}, 
namely that the same bases that pick out the dominant parts of the diagonal blocks 
(composed of the first \m{c^\ell_i} rows of the \m{Q^\ell_i} rotations) are also good for compressing 
the off-diagonal blocks. 
In the hierarchical matrix literature, for the case of specific kernels sampled in specific ways in low 
dimensions, it is possible to \emph{prove} such statements. 
In our high dimensional and less structured setting, deriving analytical results is much more challenging. 
\item MKA is an inherently bottom-up algorithm, including the clustering, thus it is naturally parallelizable 
and can be implemented in a distributed environment. 
\item The hierarchical structure of MKA is similar to that of the parallel version of MMF (pMMF)~\cite{Teneva2016}, but the way 
that the compressions are calculated is different (pMMF tries to minimize an objective that relates 
to the entire matrix).  
\end{inparaenum}

\ignore{
the form of the final approximation is 
\twocolumn[
\begin{equation}
K\approx 
C_0^\top  Q_1^\top P_1^\top\br{
C_1^\top  Q_2^\top P_2^\top \br{
X\<\oplus D_2}\ts P_2\ts Q_2\ts C_0
\<\oplus D_1}\ts P_1\ts Q_1\ts C_0
\end{equation}
]
}

\ignore{
MKA builds up a the overall matrix 

Applying compression globally to the entire kernel matrix in one stage would be very costly. 
Therefore, similary to the other 
methods described in Section \ref{sec: local}, MKA uses a fast clustering method to first block the matrix, 
and then compresses each of the diagonal blocks locally. Letting \m{C_0} denote the permutation matrix that 
rearranges the rows of \m{A} into some number, \m{p_0}, of contiguous clusters \m{K} 
is mapped to 
\[H_1=\wbar{Q}_1\,C\,A\,C^\top \wbar{Q}^\top\]
where 
\[\wbar{Q}_1=\bigoplus_i Q^1_i.\]
}
\ignore{

, and in paticular its parallelized version \cite{Teneva2016}, 
but there are some key differences. 

Similarly to other hierarchical factorization methods, MKA starts with clustering the rows/columns of 
\m{K} into some number, \m{C_0}, of clusters, blocking the matrix \m{K_0:=K} accordingly, and 
constructing an approximation to each \m{\bl{K_0}_{i,i}} diagonal block of the form 
\[\bl{K_0}_{i,i}=Q^0_i\ts H^0_i\ts {Q^0_i}^\top,\]
where \m{Q^0_i} is orthogonal. 
Also similarly to other methods, \m{Q^0_i} defines which part of the subspace defined by the \m{i}'th cluster 
(spanned by, say, the first \m{p_i} columns of \m{Q^0_i}) 
will interact with other clusters. 

In constrast 
}

\ignore{
Letting \m{A} be any \m{m\<\times m} diagonal block of \m{K}, 
the basic building blocks of MKA are what we call 
local \df{core-diagonal compressions}, which are factorizations of the form 
\begin{equation}\label{eq: compr}
A\mapsto Q^\top H\, Q=
\tikmxAb{10} 
\tikmxBb{5}{10}
\tikmxAb{10}
\end{equation}
where \m{Q} is an orthogonal matrix, and, as the figure suggests, \m{H} is \m{c}\df{--core-diagonal}, 
meaning that its non-zero entries are confined to (a) a \m{c\< \times c} block in the top left corner and 
(b) the rest of the diagonal. 
}

\section{Complexity and application to GPs}

For MKA to be effective for large scale GP regression, it must be possible to compute the factorization fast. 
In addition, the resulting approximation \m{\tilde K} must be symmetric positive semi-definite (spsd) 
(MEKA, for example, fails to fulfill this \cite{Si2014b}). 
We say that a matrix approximation algorithm \m{A\mapsto \tilde A} is \df{spsd preserving} if \m{\tilde A} 
is spsd whenever \m{A} is. It is clear from its form that the Nystr\"om approximation is spsd preserving 
, so is augmented SPCA compression.
MMF has different variants, but the core part of \m{H} is always derived 
by conjugating \m{A} by rotations, while the diagonal elements are guaranteed to be positive, therefore 
MMF is spsd preserving as well. 
\vspace{-3pt}

\begin{prop}
If the individual core-diagonal compressions in MKA are spsd preserving, then the entire algorithm is spsd perserving. 
\end{prop}
\ignore{
\begin{pf}
Assume that \m{K_{\ell-1}} is spsd. Stage \m{\ell} transforms \m{K_{\ell-1}} to \m{H_\ell=K_\ell\oplus D_\ell}. 
Here \m{K_\ell} is a submatrix of \m{\wbar{H_\ell}=\wbar{Q_\ell}\,\wbar{K_{\ell-1}}\,\wbar{Q_\ell}^\top}, therefore 
it is spsd. \m{D_\ell} is a diagonal matrix that is just the concatenation of the diagonal parts of the 
local compressions, therefore it is also spsd. By induction, if \m{K_0=K}, then all \m{K_\ell} and 
\m{D_\ell} matrices are spsd, and therefore the entire factorization \rf{eq: mka} is spsd. 
\end{pf}
}
\vspace{-3pt}

The complexity of MKA depends on the complexity of the local compressions. 
Next, we assume that to leading order in \m{m} this cost is bounded by 
\m{c_{\text{comp}}\, m^{\alpha_{\text{comp}}}} (with \m{\alpha_{\text{comp}}\<\geq 1}) 
and that each row of the \m{Q} matrix that is produced is \m{c_{\text{sp}}}--sparse. 
We assume that the MKA has \m{s} stages, 
the size of the final \m{K_s} ``core matrix'' is \m{d_{\text{core}}\times d_{\text{core}}}, 
and that the size of the largest cluster is \m{m_{\text{max}}}.
We assmue that the maximum number of clusters in any stage is \m{b_{\text{max}}} and that the 
clustering is close to balanced in the sense that 
that \m{b_{\text{max}}\<=\theta(n/m_{\text{max}})} with a small constant. 
We ignore the cost of the clustering algorithm, which varies, but usually 
scales linearly in \m{s n b_{\text{max}}}. 
We also ignore the cost of permuting the rows/columns of \m{K_\ell}, since this is a memory bound operation 
that can be virtualized away. 
The following results are to leading order in \m{m_{\text{max}}} and are similar to those 
in \cite{Teneva2016} for parallel MMF.
\vspace{-5pt}

\begin{prop}
With the above notations, the number of operations needed to compute the MKA of an \m{n\<\times n} matrix is 
upper bounded by 
\m{2 sc_{\text{sp}}n^2\<+ sc_{\text{comp}} m_{\text{max}}^{\alpha_{\text{comp}}-1}n}. 
Assuming \m{b_{\text{max}}}--fold parallelism, this complexity reduces to 
\m{2sc_{\text{sp}}n^2/b_{max}\<+ sc_{\text{comp}} m_{\text{max}}^{\alpha_{\text{comp}}}}. 
\end{prop}
\vspace{-5pt}

The memory cost of MKA is just the cost of storing the various matrices appearing in \rf{eq: mka}. 
We only include the number of non-zero reals that need to be stored and not indices, etc.. 
\vspace{-3pt}

\begin{prop}
The storage complexity of MKA is upper bounded by \m{(sc_{\text{sp}}\<+1)n+d_{\text{core}}^2}. 
\end{prop}
\vspace{-3pt}

Rather than the general case, it is more informative to 
focus on MMF based MKA, which is what we use in our experiments. 
We consider the simplest case of MMF, referred to as ``greedy-Jacobi'' MMF, in 
which each of the \m{q_i} elementary rotations is a Given rotation. An additional parameter of 
this algorithm is the compression ratio \m{\gamma}, which in our notation is equal to \m{c/n}. 
Some of the special features of this type of core-diagonal compression are: 
\begin{compactenum}[(a)]
\item While any given row of the rotation \m{Q} produced by the algorithm is not guaranteed to 
be sparse, \m{Q} will be the product of exactly \m{\lfloor (1\<-\gamma) m \rfloor} Givens rotations. 
\item The leading term in the cost is the \m{m^3} cost of computing \m{A^\top\! A}, 
but this is a BLAS operation, so it is fast. 
\item Once \m{A^\top\! A} has been computed, the cost of the rest of the compression scales with \m{m^2}.  
\end{compactenum}
Together, these features result in very fast core-diagonal compressions and a very compact 
representation of the kernel matrix. 
\vspace{-5pt}

\begin{prop}
The complexity of computing the MMF-based MKA of an \m{n\<\times n} dense matrix is upper bounded by 
\m{4sn^2+sm_{\text{max}}^2 n}, where \m{s=\log(d_\text{core}/n)/(\log \gamma)}. 
Assuming \m{b_{\text{max}}}--fold parallelism, this is reduced to  
\m{4sn m_{\text{max}}+m_{\text{max}}^3}.    
\end{prop}
\vspace{-5pt}

\begin{prop}
The storage complexity of MMF-based MKA is upper bounded by \m{(2s\<+1)n+d_{\text{core}}^2}. 
\end{prop}
\vspace{-5pt} 

Typically, \m{d_{\text{core}}=O(1)}. Note that this implies \m{O(n\log n)} storage complexity, 
which is similar to Nystr\"om approximations with very low rank. Finally, we have the following 
results that are critical for using MKA in GPs.\vspace{-5pt}

\begin{prop}\label{prop: mka mult}
Given an approximate kernel \m{\tilde K} in MMF-based MKA form \rf{eq: mka}, and a vector \m{\V z\tin\RR^n}  
the product \m{\tilde K\V z} can be computed in \m{4sn\<+d_{\text{core}}^2} operations. 
With \m{b_{\text{max}}}--fold parallelism, this is reduced to \m{4sm_{\text{max}}+d_{\text{core}}^2}. 
\end{prop}
\vspace{-5pt}

\begin{prop}\label{prop: mka pow}
Given an approximate kernel \m{\tilde K} in (MMF or SPCA-based) MKA form, the MKA form of \m{\tilde K^\alpha} 
for any \m{\alpha} can be computed in \m{O(n+d_{\text{core}}^3)} operations. 
The complexity of computing the matrix exponential \m{\exp(\beta\tilde K)} for any \m{\beta} 
in MKA form and the complexity of computing \m{\det(\tilde K)} are also \m{O(n+d_{\text{core}}^3)}. 
\end{prop}
\vspace{-5pt}

\subsection{MKA--GPs and MKA Ridge Regression}

The most direct way of applying MKA to speed up GP regression (or ridge regression) 
is simply using it to approximate the augmented kernel matrix \m{K'=(K\<+\sigma^2 I)} 
and then inverting this approximation using Proposition \ref{prop: mka pow} (with \m{\alpha\<=-1}). 
Note that the resulting \m{\tilde K'{}^{-1}} never needs to be evaluated fully, in matrix form. 
Instead, in equations such as \rf{eq: gp map}, the matrix-vector product \m{\tilde K'{}^{-1}\V y} 
can be computed in ``matrix-free'' form by cascading \m{\V y} through the analog of \rf{eq: mka}. 
Assuming that \m{d_{\text{core}}\ll n} and \m{m_{\text{max}}} is not too large, 
the serial complexity of each stage of this computation scales with at most \m{n^2}, 
which is the same as the complexity of computing \m{K} in the first place.  

One potential issue with the above approach however is that 
because MKA involves repeated truncation of the \m{H_j^{\text{pre}}} matrices, 
\m{\tilde K'} will be a biased approximation to \m{K},  
therefore expressions such as \rf{eq: gp map} which mix an approximate \m{K'} with an exact 
\m{\V k_x} will exhibit some systematic bias. 
In Nystr\"om type methods (specifically, the so-called Subset of Regressors and Deterministic 
Training Conditional GP approximations) this problem is addressed by replacing 
\m{\V k_x} with its own Nystr\"om approximation, \m{\hat {\V k}_x\<=K_{\ast,I} W^{+} {\V k}_x^{I}}, 
where \m{[\hat k_x^{I}]_j\<=k(x,x_{i_{j}})}. 
Although \m{\hat K'=K_{\ast,I} W^{+} K_{\ast,I}^\top\<+\sigma^2 I} is a large matrix, expressions 
such as \m{\hat {\V k}{}_x^\top \hat K'{}^{-1}} can nonetheless be efficiently evaluated by using a 
variant of the Sherman--Morrison--Woodbury identity and the fact that \m{W} 
is low rank (see \cite{CandelaR05}). 

The same approach cannot be applied to MKA because \m{\tilde K} is not low rank.  
Assuming that the testing set \m{\cbrN{\sseq{x}{p}}} is known at training time, however,
instead of approximating \m{K} or \m{K'}, we compute the MKA approximation of 
the joint train/test kernel matrix 
\vspace{-5pt}
\begin{equation*}\label{eq: jointK}
\Kcal=\br{
\begin{array}{c|c}
K&K_{\ast}\\
\hline 
K_\ast^\top&K_{\text{test}}\\
\end{array}}
\qquad\text{where}\qquad 
\begin{array}{ll}
&K_{i,j}=k(x_i,x_j)+\sigma^2\\
&[K_{\ast}]_{i,j}=k(x_i,x'_j)\\
&[K_{\text{test}}]_{i,j}=k(x'_i,x'_j).\\
\end{array}\vspace{-5pt}
\end{equation*}
Writing \m{\Kcal^{-1}} in blocked form 
\[
\tilde \Kcal^{-1}=
\br{\begin{array}{c|c}
A&B\\
\hline
C&D\\ 
\end{array}},
\]
and taking the Schur complement of \m{D} now recovers an alternative approximation 
\m{\check{K}^{-1}=A-BD^{-1} C}  
to \m{K^{-1}} which is consistent with the off-diagonal block \m{K^{\ast}} leading to our final 
MKA--GP formula  
\m{\V{\h f}=K_{\ast}^{\top} \check{K}^{-1} \V y},
where \m{\V{\h f}=(\h f(x'_1),\ldots,\h f(x'_p))^\top}. 
While conceptually this is somewhat more involved than naively estimating \m{K'}, assuming 
\m{p\ll n}, the cost of inverting \m{D} is negligible, and the overall serial complexity of the 
algorithm remains \m{(n+p)^2}. 


\ignore{
\subsection{Fast GP Regression from Explicit \m{K}}

The above propositions are all that are needed to use MKA for fast Gaussian Process prediction, 
and the results in the experiments section show that MKA is very adept at capturing structure 
in kernel matrices \emph{across the entire spectrum} from low eigenvalues to high. 
In particular, since Proposition \ref{prop: mka mult} affords \m{O(n\log n)} matrix/vector multiplies, 
it can be used inside of an iterative method for solving \m{\tilde K' \V z\<=\V b}. 
Even more to the point, in conjunction with Proposition \ref{prop: mka pow}, it can be used to 
approximate each of the expressions 
\m{\V k_x {K'}^{-1} \V y}, \m{\V k_x {K'}^{-1} \V k_{x'}} and 
\m{\det(K')} appearing in \rf{eq: gp map}, \rf{eq: gp pvar} and \rf{eq: gp ll} directly, 
also in \m{O(n\log n)} time. 
Thus, the complexity of the entire GP prediction process is dominated by the \m{O(n^2)} complexity 
of computing the factorization itself, which is the same as the complexity of computing \m{K} in the first place. 

One potential issue, however, is that \m{\tilde K'} is not an unbiased estimator of \m{K'} 
(specifically because MMF explicit truncation to core-diagonal form), 
therefore expressions such as \m{\V k_x \tilde K} have systematic errors. 
To compensate for this, assuming that in addition to the training set \m{\cbr{\sseq{x}{n}}}, the test set 
\m{\cbr{\sseq{x}{p}}} is also known in advance, we compute the MKA of the joint train/test kernel matrix 
\vspace{-5pt}
\begin{equation*}\label{eq: jointK}
\Kcal=\br{
\begin{array}{c|c}
K&K_{\ast}\\
\hline 
K_\ast^\top&K_{\text{test}}\\
\end{array}
}
\quad\text{where}
\begin{array}{ll}
&K_{i,j}=k(x_i,x_j)\\
&[K_{\ast}]_{i,j}=k(x_i,x'_j)\\
&[K_{\text{test}}]_{i,j}=k(x'_i,x'_j).\\
\end{array}\vspace{-5pt}
\end{equation*}
By taking the Schur complement of \m{D} in 
\[
\tilde \Kcal^{-1}=
\br{\begin{array}{c|c}
A&B\\
\hline
C&D\\ 
\end{array}}
\]
we get the approximation \m{K^{-1}\approx A-BD^{-1} C}. 
For \m{p\<\ll n}, the \m{O(p^3)}cost of computing \m{D^{-1}} is negligible. 
However, since \m{\tilde \Kcal} is typically not expressed explicitly in matrix form, 
the cost of extracting \m{D} from the factorized form of \m{\Kcal} is \m{O(p(n\<+p)\log(n\<+p))}. 
}


In certain GP applications, the \m{O(n^2)} cost of writing down the kernel matrix 
is already forbidding. 
The one circumstance under which MKA can get around this problem is 
when the kernel matrix is a matrix polynomial in a sparse 
matrix \m{L}, which is most notably for diffusion kernels and certain other graph kernels. 
Specifically in the case of MMF-based MKA, 
since the computational cost is dominated by computing local ``Gram matrices'' \m{A^\top\! A},
when \m{L} is sparse, and this sparsity is retained from one compression to another, the MKA of sparse  
matrices can be computed very fast. 
In the case of graph Laplacians, empirically, the complexity is close to linear in \m{n}. 
By Proposition \ref{prop: mka pow}, the diffusion kernel and certain other graph kernels can also be approximated 
in about \m{O(n\log n)} time. 

\ignore{
\subsection{Combined MKA/Nystr\"om method}

The main reason that in Nystr\"om type methods the rank \m{k} is usually taken to be small is that 
the \m{k\<\times k} matrix \m{W} needs to be inverted, which is an expensive operation. 
It is then natural to combine the advantages of Nystr\"om and MKA, by sampling a large number 
of rows/columns from \m{K} (say, on the order of \m{n/10}) and then using MKA to compute \m{W^+}. 
Unfortunately, space limitations prevent us from further expanding on this method. 
}

\ignore{
As explained in \ref{sec: disc}, while not the only option,  
the simplest choice is to use a ``mini multiresolution factorization'' for this purpose, specifically, 
an MMF algorithm. 
Both MMF and the present work are inspired by the structure of fast wavelet transforms 
\cite{Mallat1989}, in fact, superficially, MMF has a very similar form to MKA, 
\begin{equation*}\label{eq: mmf}
A\approx q_1^\top \ldots q_L^\top H\, q_L\ldots q_1.
\end{equation*}
In MMF, however, the \m{q_i}'s are elementary rotations (in the simplest case, Givens rotations),  
and of course there is no notion of blocking or stages involved, instead,  
the \m{q_i}'s are determined by explicitly minimizing an objective function related to the 
Frobenius norm error. 
Assuming that \m{A\tin\RR^{m\times m}}, the cost of computing an (unblocked) dense MMF is 
dominated by the \m{m^3} complexity of computing \m{A^2}, 
which will be a strong reason to keep the size of the clusters in each stage of MKA small. 
The number of rotations involved in the factorization, however, is only \m{O(m)}, so storing 
and applying an existing MMF to vectors is very efficient. 
\begin{prop} (c.f., \cite{MMFicml2014}) 
The 
\end{prop}

Another separate component of MKA is the clustering algorithm used in each stage. In practice, 
clustering is not the dominant factor in complexity, especially since the clustering can be quite rough. 
In our experiments, the block \m{K_s} into \m{p_s\times p_s} clusters, 
we use a fast inner product based clustering algorithm that selects \m{O(p_s)} columns of \m{K_s}  
uniformly at random to be anchors, 
assigns each other column to one of the anchors based on normalized inner product, and has an iterative 
refinement loop that merges clusters that are too small and splits clusters that are too large. 

\begin{prop}

\end{prop}
}

\section{Experiments}
\begin{figure}[t!]
	\centering
	\begin{minipage}[b]{0.16\linewidth}
		\centering \tiny{\textbf{Full}}\\
		\includegraphics[width=1\textwidth]{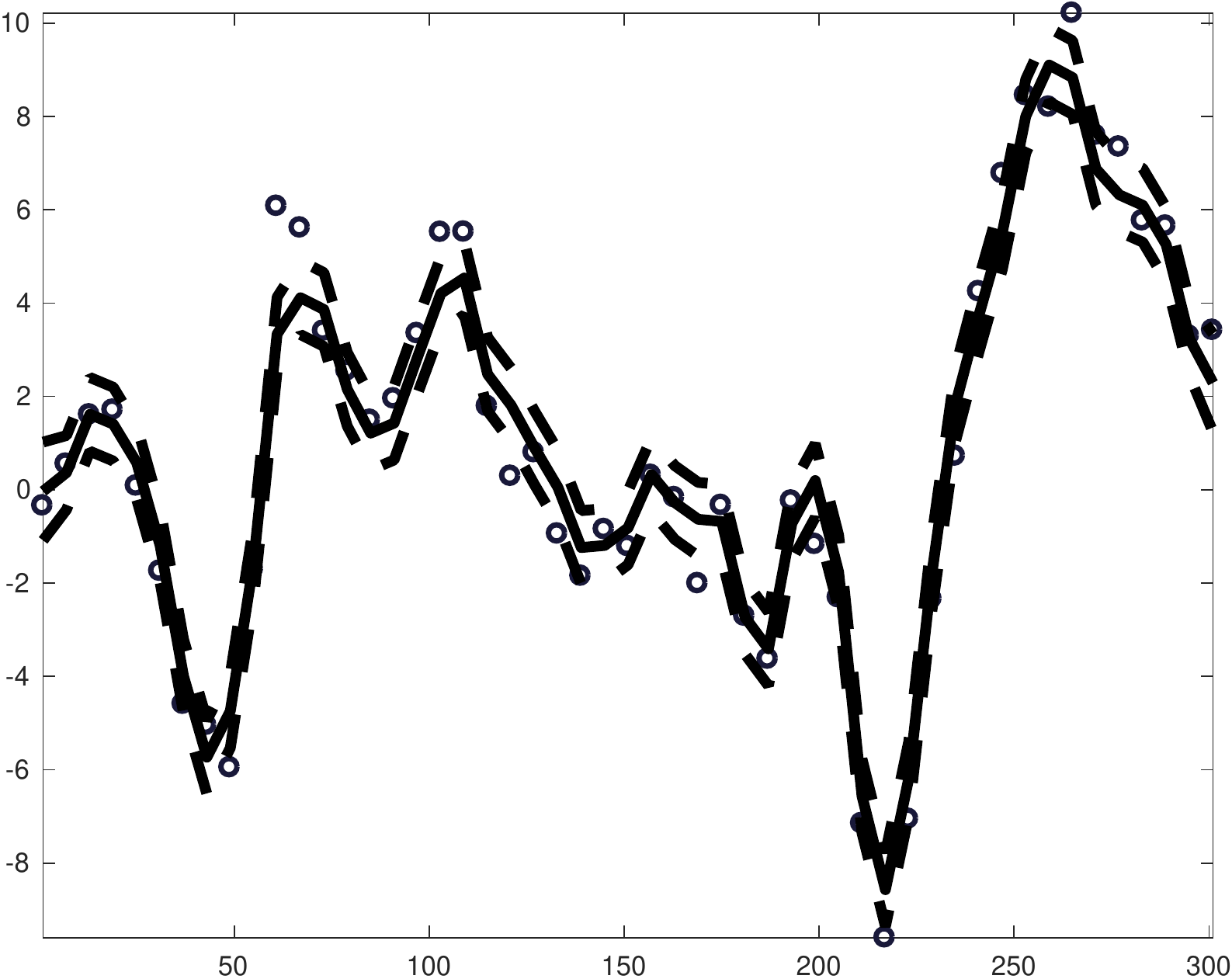}  
	\end{minipage}
	\begin{minipage}[b]{0.16\linewidth}
		\centering \tiny{\textbf{SOR}}\\
\includegraphics[width=1\textwidth]{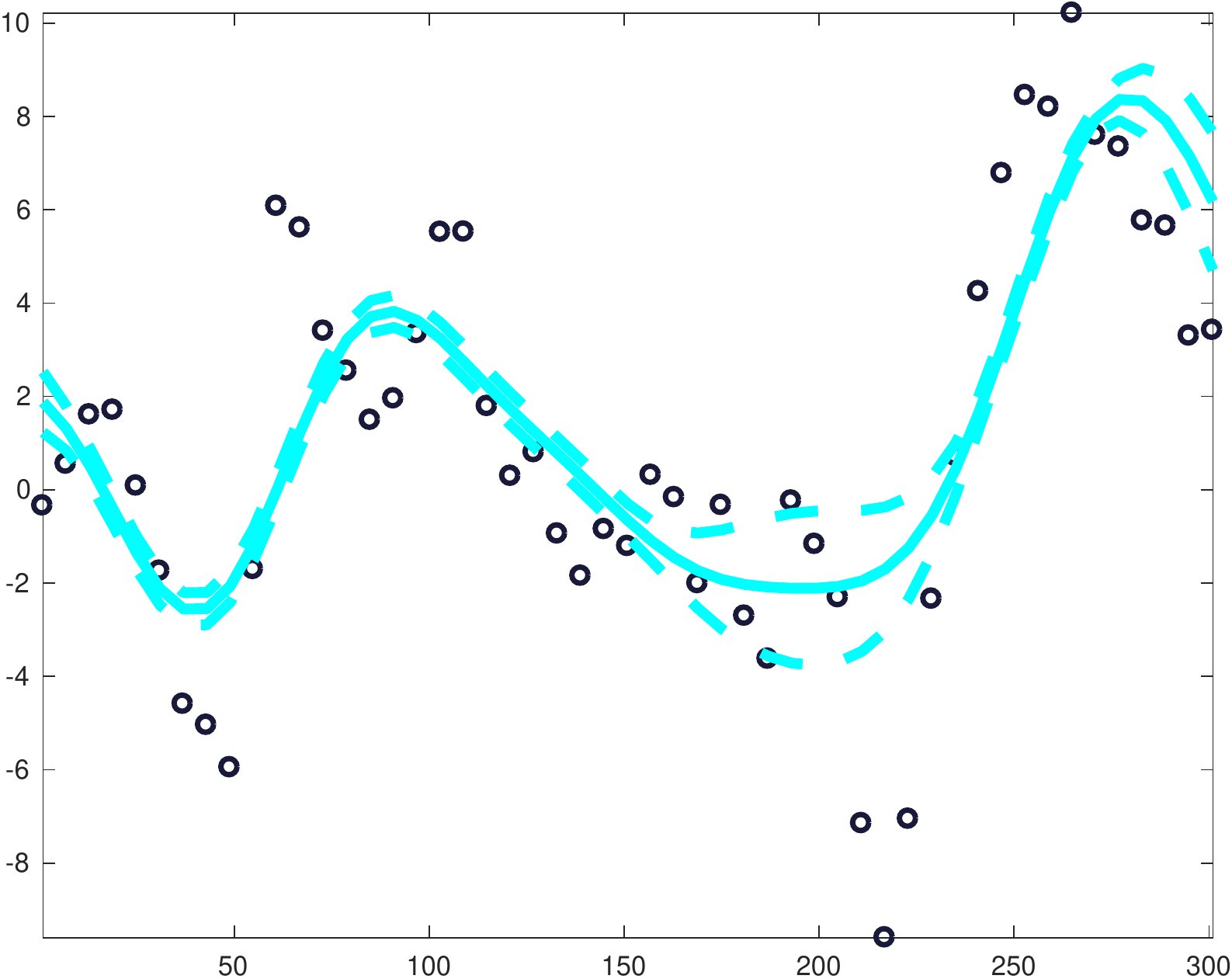}
	\end{minipage}
	\vspace{1pt}
	\begin{minipage}[b]{0.16\linewidth}
		\centering \tiny{\textbf{FITC}}\\
\includegraphics[width=1\textwidth]{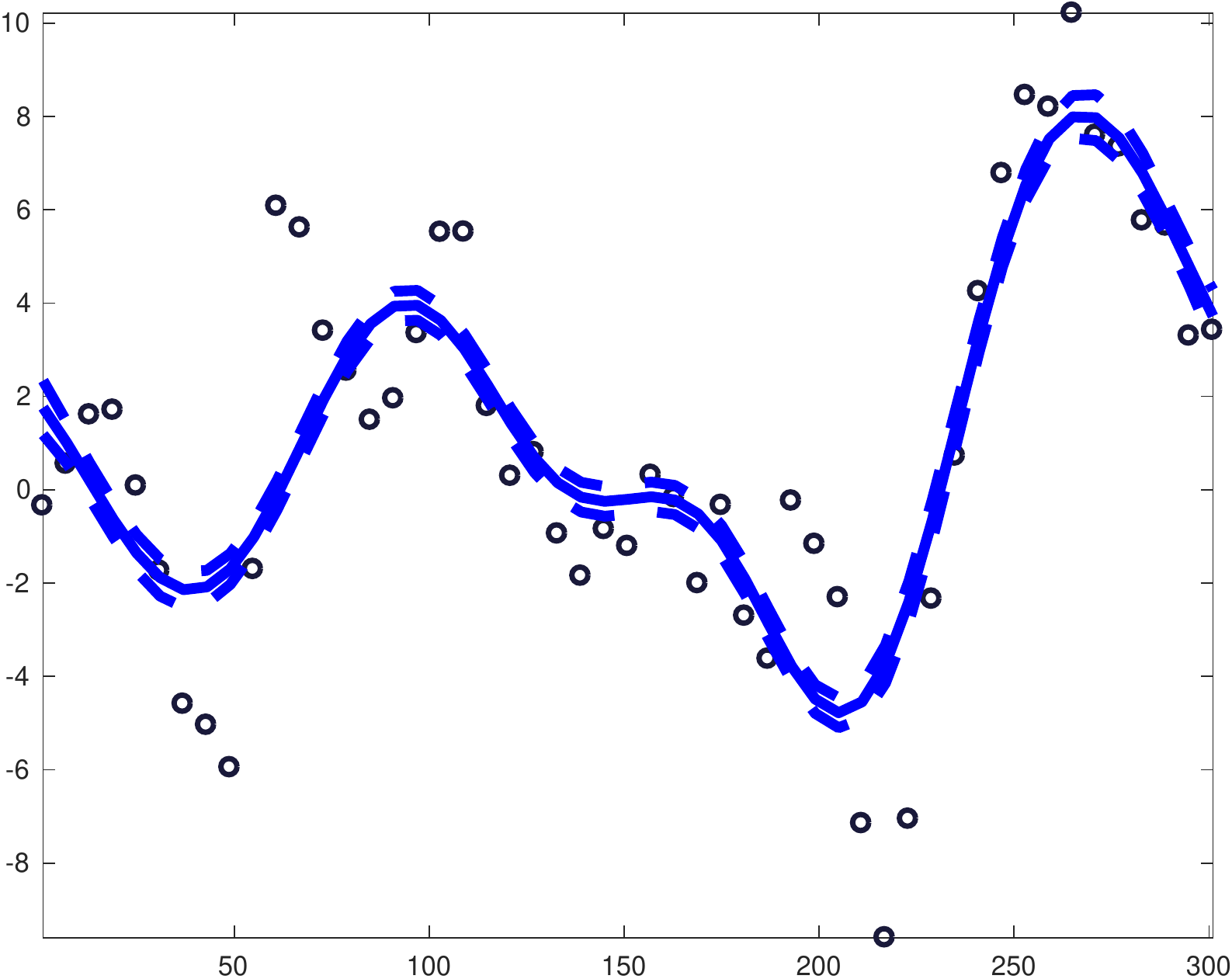}
	\end{minipage} 
	\begin{minipage}[b]{0.16\linewidth}
		\centering \tiny{\textbf{PITC}}\\ 
\includegraphics[width=1\textwidth]{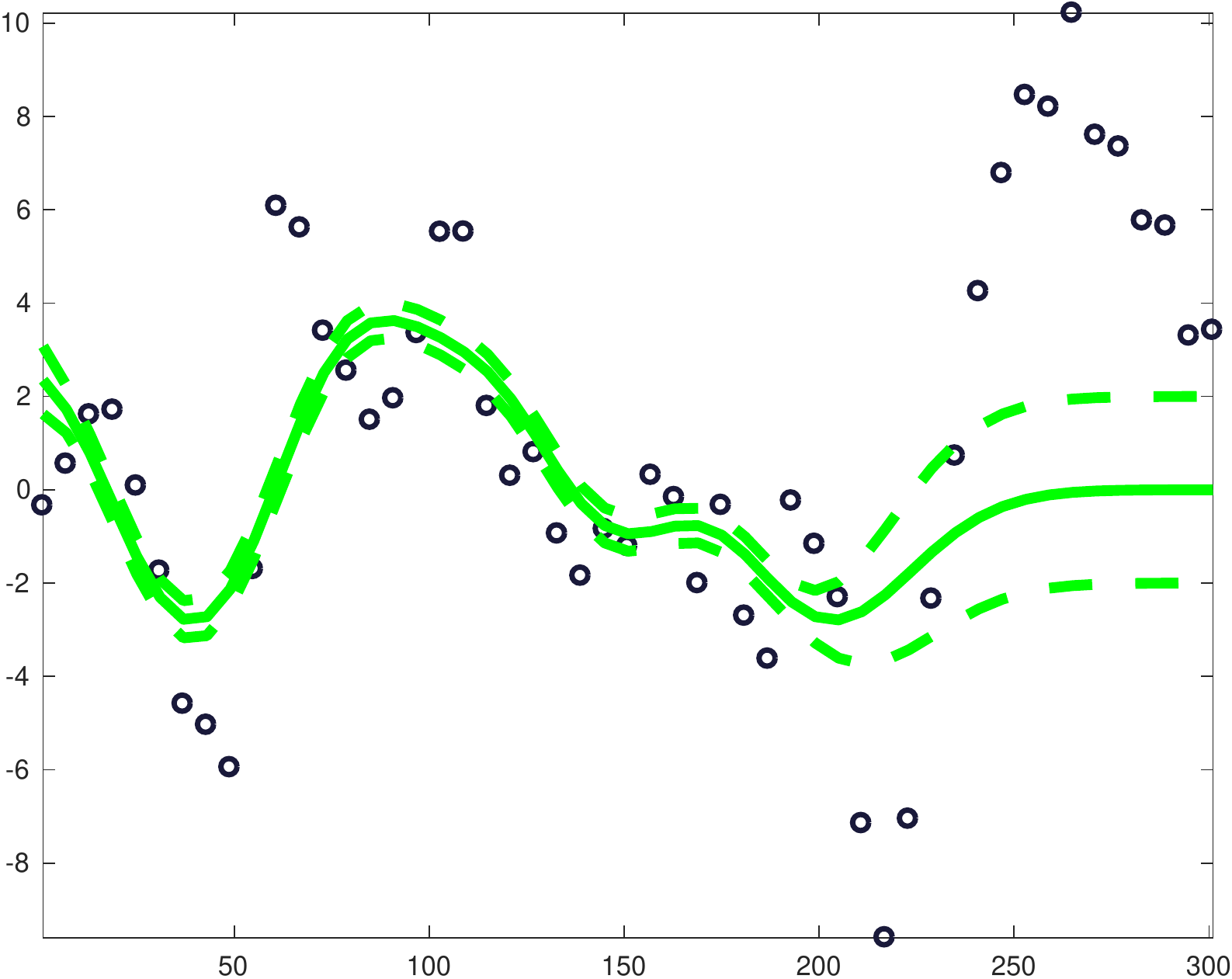}
	\end{minipage}
	\vspace{1pt}
	\begin{minipage}[b]{0.16\linewidth}
		\centering \tiny{\textbf{MEKA}}\\ 
\includegraphics[width=1\textwidth]{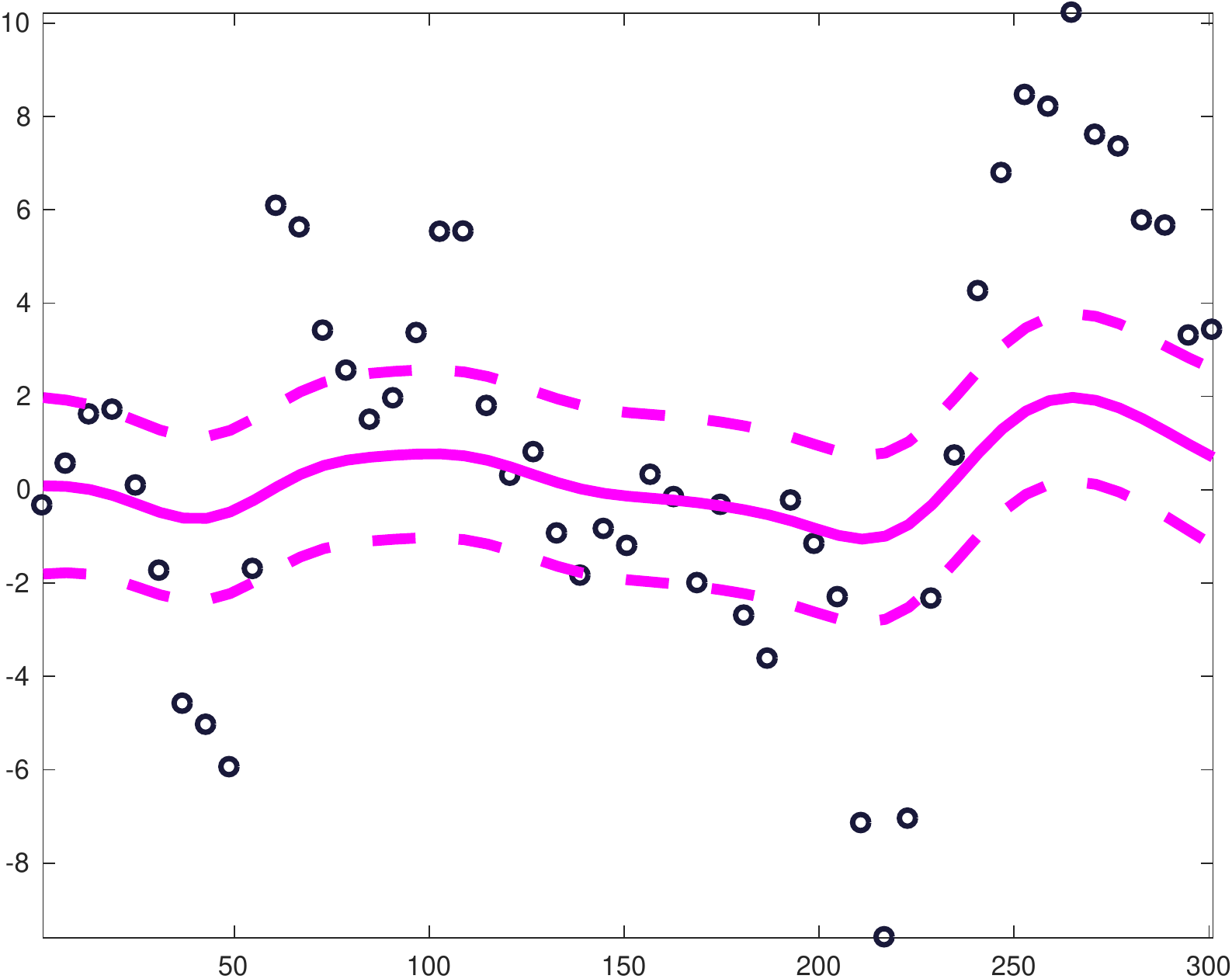}
	\end{minipage}
	\begin{minipage}[b]{0.16\linewidth}
		\centering \tiny{\textbf{MKA}}\\ 
\includegraphics[width=1\textwidth]{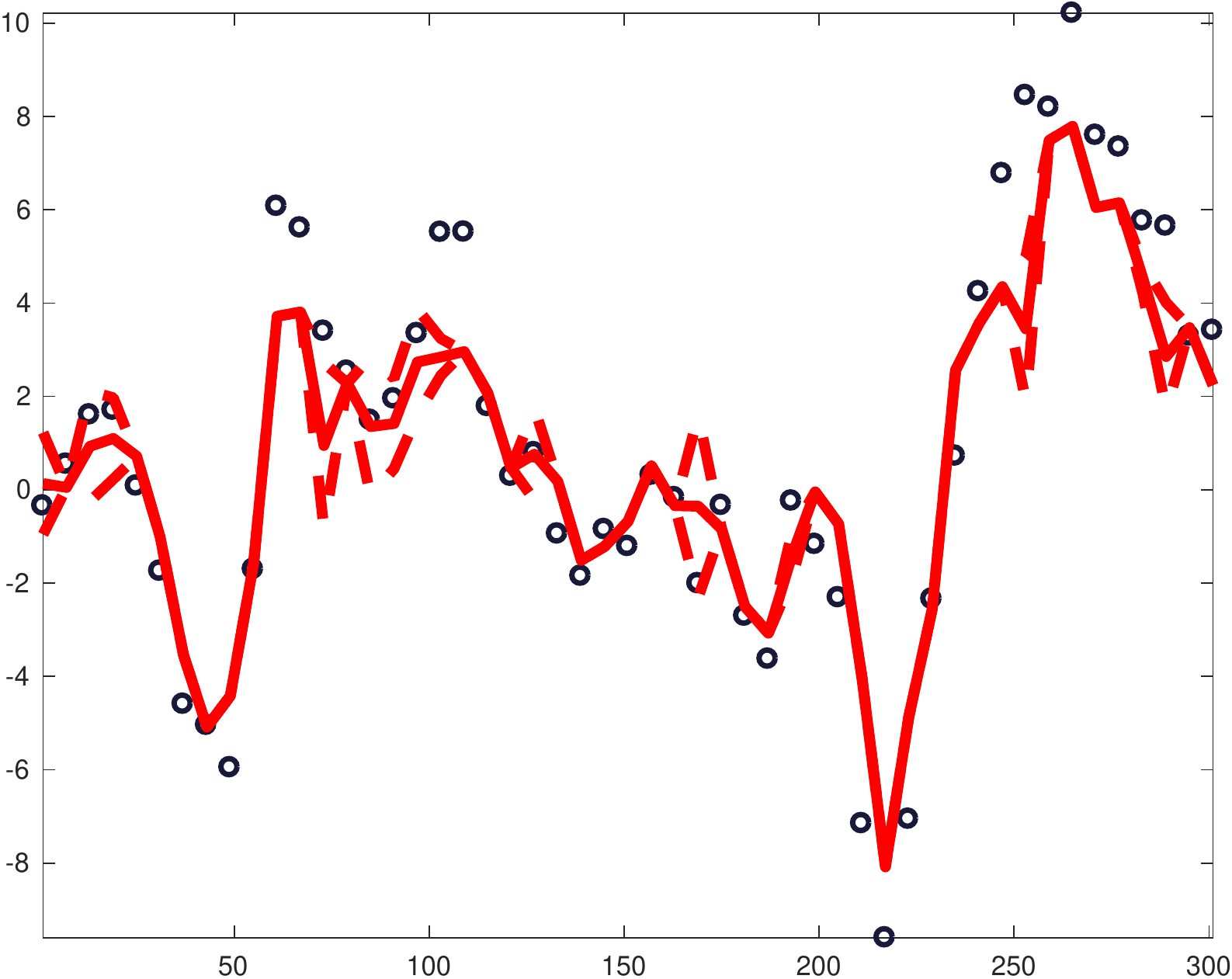}
	\end{minipage} 
	\vspace{-10pt}
	\caption{Snelson's 1D example: ground truth (black circles); prediction mean (solid line curves); one standard deviation in prediction uncertainty (dashed line curves).}\label{fig:toy}  
\end{figure}
\begin{table*}[]
\small
\centering
\vspace{-10pt}
\caption{\label{tbl: res1}Regression Results with $k$ to be $\#$ pseudo-inputs/\m{d_{\text{core}}} :  SMSE(MNLP)}
\begin{adjustbox}{width=1\textwidth}
\begin{tabular}{ l c c c c c c c}
\toprule
Method     &k  & Full & SOR  & FITC & PITC & MEKA & MKA \\
\hline
housing    &16 &\m{0.36(-0.32)} & \m{0.93( -0.03)} & \m{0.91(-0.04)} & \m{0.96( -0.02)} & \m{0.85( -0.08)}& \boldmath{\m{0.52( -0.32)}}  \\
rupture    &16 &\m{0.17(-0.89)} & \m{0.94( -0.04)} & \m{0.96(-0.04)} & \m{0.93( -0.05)} & \m{0.46( -0.18)}& \boldmath{\m{0.32( -0.54)}}  \\
wine       &32 &\m{0.59(-0.33)} & \m{0.86( -0.07)} & \m{0.84(-0.03)} & \m{0.87( -0.07)} & \m{0.97( -0.12)}& \boldmath{\m{0.70( -0.23)}}  \\
pageblocks &32 &\m{0.44(-1.10)} & \m{0.86( -0.57)} & \m{0.81(-0.78)} & \m{0.86( -0.72)} & \m{0.96( -0.10)}& \boldmath{\m{0.63( -0.85)}}  \\ 
compAct    &32 &\m{0.58(-0.66)} & \m{0.88( -0.13)} & \m{0.91(-0.08)} & \m{0.88( -0.14)} & \m{0.75( -0.21)}& \boldmath{\m{0.60( -0.32)}}  \\
pendigit   &64 &\m{0.15(-0.73)} & \m{0.65( -0.19)} & \m{0.70(-0.17)} & \m{0.71( -0.17)} & \m{0.53( -0.29)}& \boldmath{\m{0.30( -0.42)}}  \\ 
\bottomrule
\end{tabular}
\end{adjustbox}
\vspace{-10pt}
\end{table*}

\begin{figure}[!htbp]
	\centering
	\begin{minipage}[b]{0.24\linewidth}
		\centering \tiny{\textbf{housing}}\\
		\includegraphics[width=1\textwidth]{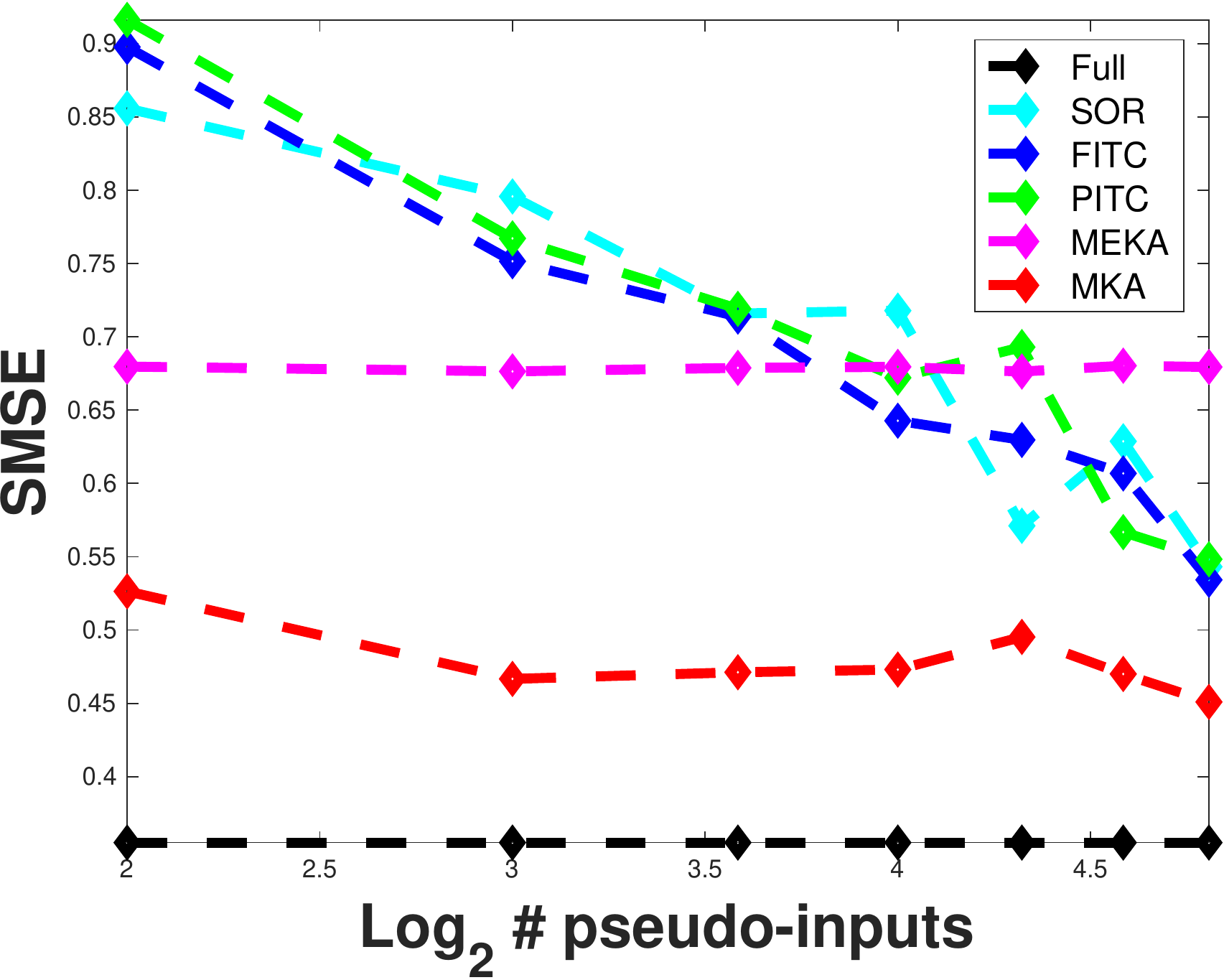}
	\end{minipage}
	\begin{minipage}[b]{0.24\linewidth}
		\centering \tiny{\textbf{housing}}\\
        \includegraphics[width=1\textwidth]{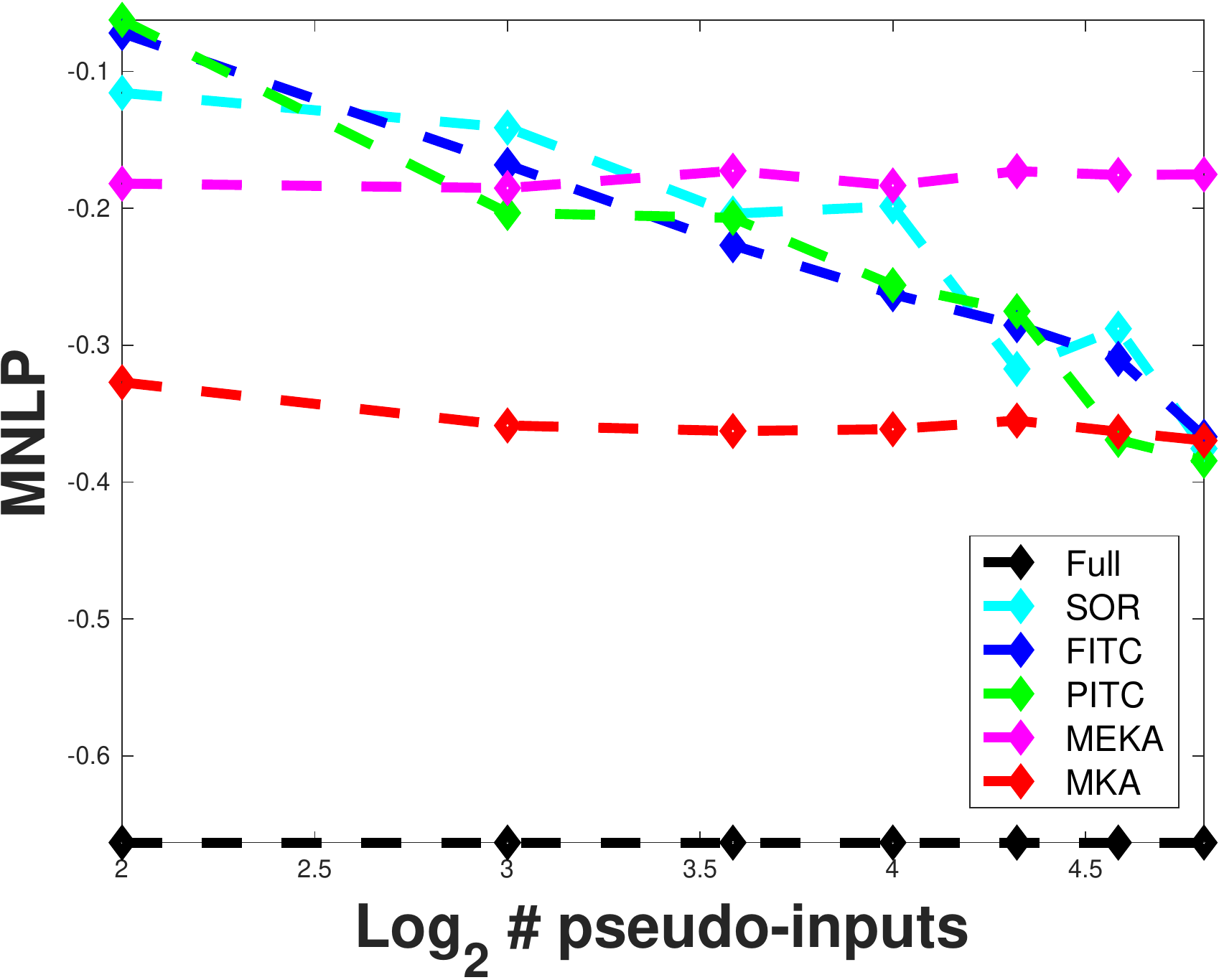}
	\end{minipage} 
	\begin{minipage}[b]{0.24\linewidth}
		\centering \tiny{\textbf{rupture}}\\
	\includegraphics[width=1\textwidth]{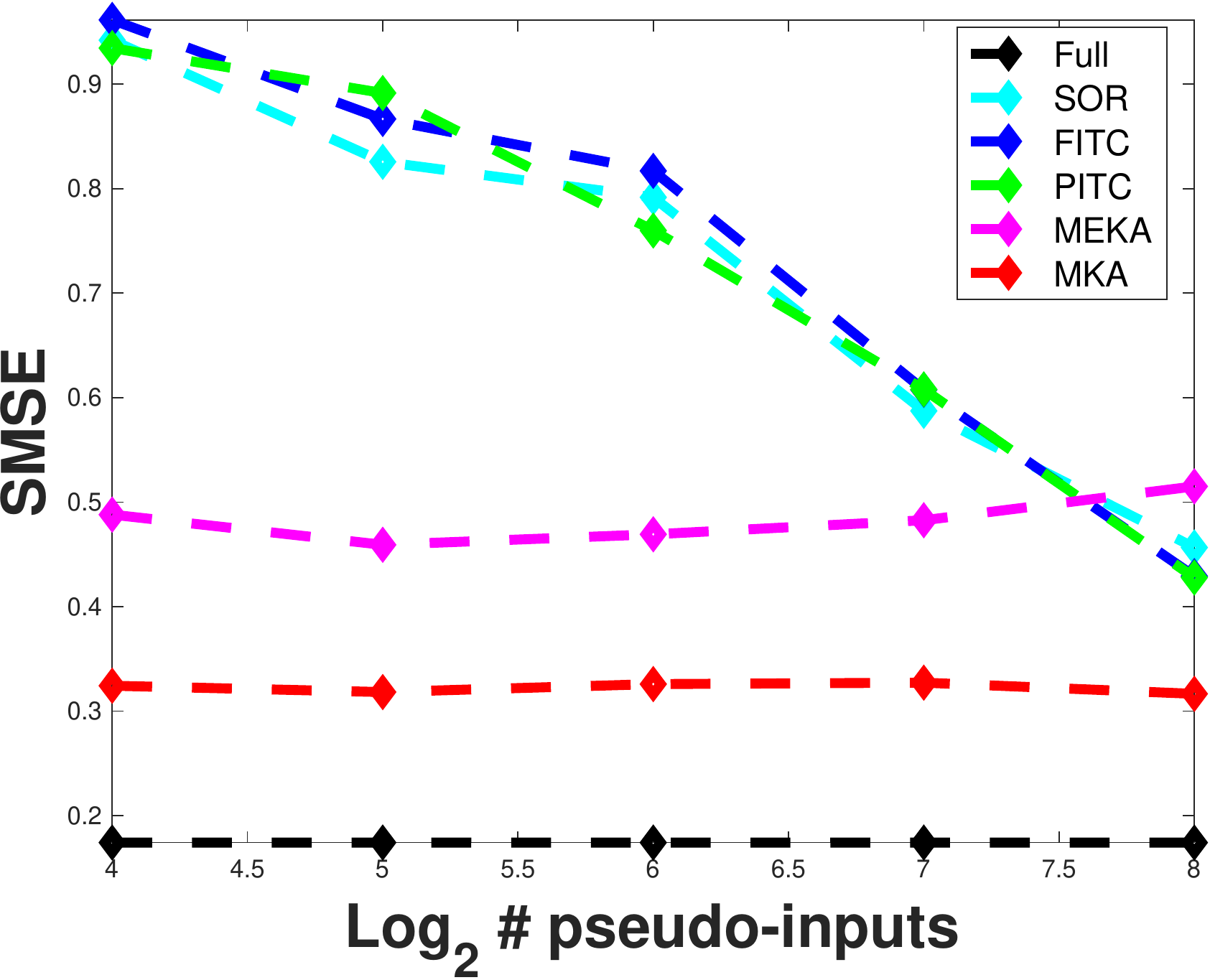}
	\end{minipage}
	\begin{minipage}[b]{0.24\linewidth}
		\centering \tiny{\textbf{rupture}}\\
	\includegraphics[width=1\textwidth]{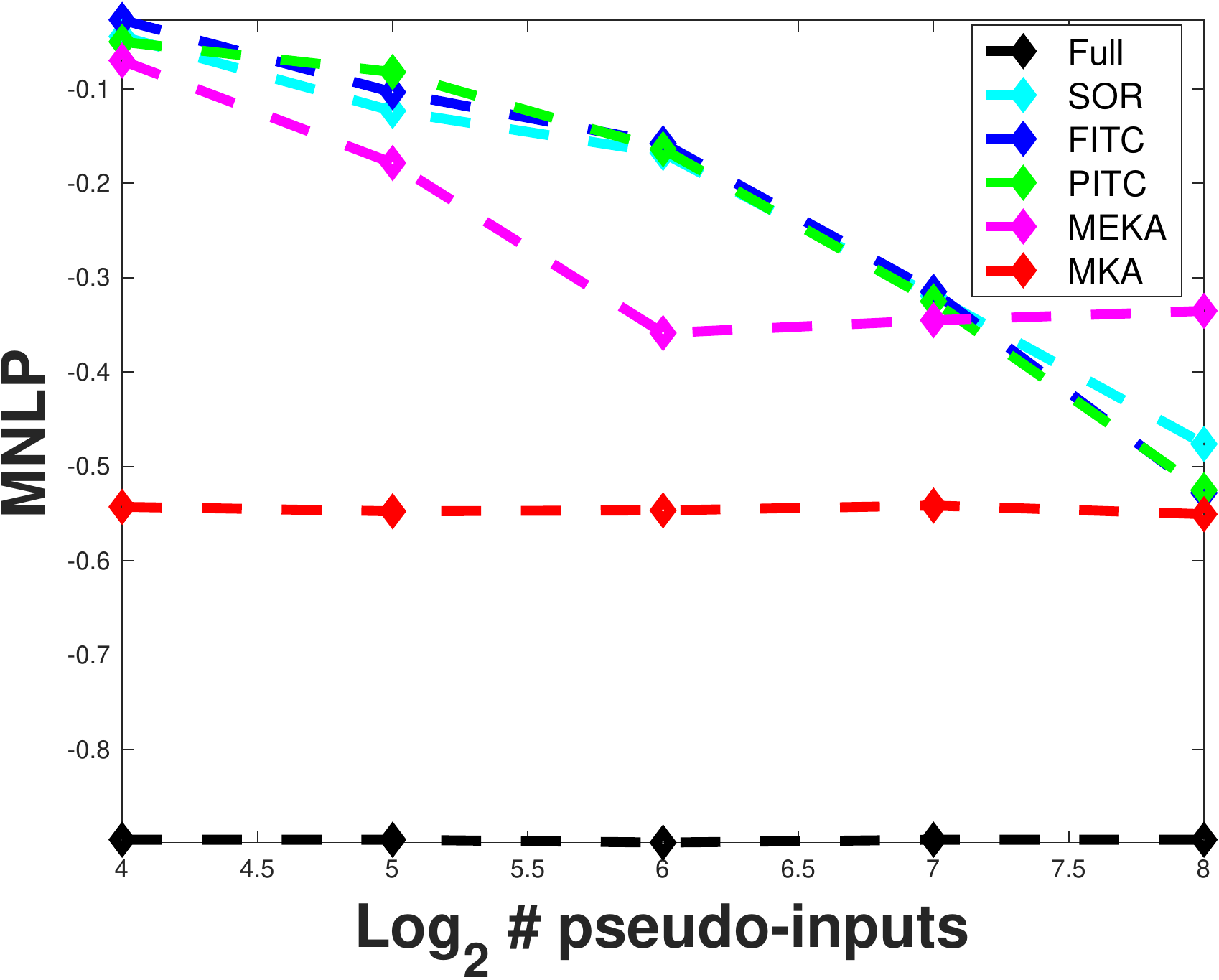}
	\end{minipage}
	\vspace{-8pt}
	\caption{SMSE and MNLP as a function of the number of pseudo-inputs/\m{d_{\text{core}}} on two datasets. In the given range MKA clearly outperforms the other methods in both error measures.
	}\label{fig:rank-main}  \vspace{-0.1in}
\end{figure}

We compare MKA to five other methods: 
\begin{inparaenum}[~1.]
\item \textbf{Full}: the full GP regression using Cholesky factorization~\cite{RasmussenBook}. 
\item \textbf{SOR}: the Subset of Regressors method 
(also equivalent to DTC in mean)~\cite{RasmussenBook}.  
\item \textbf{FITC}: the Fully Independent Training Conditional approximation, 
also called Sparse Gaussian Processes using Pseudo-inputs~\cite{SnelsonG05}. 
\item \textbf{PITC}: the Partially Independent Training Conditional approximation method (also equivalent to PTC in mean)~\cite{CandelaR05}.
\item \textbf{MEKA}: the Memory Efficient Kernel Approximation method~\cite{Si2014b}. 
\end{inparaenum}
The KISS-GP~\cite{WilsonN15} and other interpolation based methods are not discussed in this paper, because, we believe, they mostly only apply to low dimensional settings. We used custom Matlab implementations~\cite{RasmussenBook} for Full, SOR, FITC, and PITC. 
We used the Matlab codes provided by the author for MEKA. 
Our algorithm MKA was implemented in C++ with the Matlab interface. 
To get an approximately fair comparison, we set \m{d_{\text{core}}} in MKA to be the number of pseudo-inputs. 
The parallel MMF algorithm was used as the compressor due to its
computational strength~\cite{Teneva2016}.
The Gaussian kernel is used for all experiments with one length scale for all input dimensions. 

\textbf{Qualitative results.}
We show the qualitative behavior of each method on the 1D toy dataset from \cite{SnelsonG05}. 
We sampled the ground truth from a Gaussian processes with length scale \m{\ell=0.5} and number of 
pseudo-inputs (\m{d_{\text{core}}}) is 10. 
We applied cross-validation to select the parameters for each method to fit the data. 
Figure~\ref{fig:toy} shows that MKA fits the data almost as well as the Full GP does. 
In terms of the other approximate methods, although their fit to the data is smoother, 
this is to the detriment of capturing the local structure of the underlying data, 
which verifies MKA's ability to capture the entire spectrum of the kernel matrix, 
not just its top eigenvectors.

\textbf{Real data.} 
We tested the efficacy of GP regression on real-world datasets.
The data are normalized to mean zero and variance one.  
We randomly selected 10\% of each dataset to be used as a test set. 
On the other 90\% we did five-fold cross validation to learn the length scale and noise parameter 
for each method and the regression results were averaged over repeating this setting five times. 
All experiments were ran on a 3.4GHz 8 core machine with 8GB of memory. 
Two distinct error measures are used to assess performance:  (a) standardized mean square error (SMSE), 
$\frac{1}{n}\sum_{t=1}^{n}(\hat{y}_t-y_t)^2/\hat{\sigma}^2_{\star}$, 
where $\hat{\sigma}^2_{\star}$ is the variance of test outputs,  
and (2) mean negative log probability (MNLP) 
$\frac{1}{n}\sum_{t=1}^{n}\left((\hat{y}_t-y_t)^2/\hat{\sigma}^2_{\star}+\log \hat{\sigma}^2_{\star} + \log 2\pi\right)$, 
each of which corresponds to the predictive mean and variance in error assessment. 
From Table~\ref{tbl: res1}, we are competitive in both error measures when the number of 
pseudo-inputs (\m{d_{\text{core}}}) is small, 
which reveals low-rank methods' inability in capturing the local structure of the data.
We also illustrate the performance sensitivity by varying the number of pseudo-inputs on selected datasets. 
In Figure~\ref{fig:rank-main}, for the interval of pseudo-inputs considered, MKA's performance is
robust to \m{d_{\text{core}}}, while low-rank based methods' performance changes rapidly, which shows
MKA's ability to achieve good regression results even with a crucial compression level.
The Supplementary Material gives a more detailed discussion of the datasets and experiments.

\section{Conclusions}

In this paper we made the case that whether a learning problem 
is low rank or not depends on the nature of the data rather than just the spectral properties of the kernel matrix \m{K}. 
This is easiest to see in the case of Gaussian Processes, which is the algorithm that we focused on in this paper, but it is also true more generally. 
Most existing sketching algorithms used in GP regression force low rank structure on \m{K}, either globally, or at the block level. When the nature of the problem is indeed low rank, this might 
actually act as an additional regularizer and improve performance. 
When the data does not have low rank structure, however, low rank approximations will fail. 
Inspired by recent work on multiresolution factorizations, 
we proposed a mulitresolution meta-algorithm, MKA, for approximating kernel matrices, 
which assumes that the \emph{interaction} between distant clusters is low rank, while avoiding forcing 
a low rank structure of the data locally, at any scale.  
Importantly, MKA allows fast direct calculations of the inverse of the kernel matrix and its determinant, 
which are almost always the computational bottlenecks in GP problems. 


\subsection*{Acknowledgements}\label{sec: acknowledgements}
This work was completed in part with resources provided
by the University of Chicago Research Computing Center. 
The authors wish to thank Michael Stein for helpful suggestions. 


\clearpage

{\small
\setlength{\bibsep}{0pt}
\bibliographystyle{plain} 
\bibliography{mgp}
}

\end{document}


\maketitle
\vspace{-10pt}
	
\section{Block structure for different hierarchical matrix approximations}

\newcommand{\tikquad}[4]{
\draw (0+#3,#1+#4) rectangle +(#1,#1);
\draw (#1+#3,0+#4) rectangle +(#1,#1);
\pgfmathtruncatemacro{\truncr}{#2}
\ifthenelse{\truncr=0}
{\filldraw[gray](0+#3,#1+#4) rectangle +(#1,#1); \filldraw[gray](#1+#3,0+#4) rectangle +(#1,#1);}
{\tikquad{#1/2}{#2-1}{#3}{#4+#1}\tikquad{#1/2}{#2-1}{#3+#1}{#4}}
}

\newcommand{\tikHa}[4]{
\draw (0+#3,#1+#4) rectangle +(#1,#1);
\draw (#1+#3,0+#4) rectangle +(#1,#1);
\pgfmathtruncatemacro{\truncr}{#2}
\ifthenelse{\truncr=0}
{\filldraw[gray](0+#3,#1+#4) rectangle +(#1,#1); \filldraw[gray](#1+#3,0+#4) rectangle +(#1,#1);}
{
\tikHa{#1/2}{#2-1}{#3}{#4+#1}
\tikHa{#1/2}{#2-1}{#3+#1}{#4}
\tikHb{#1/2}{#2-1}{#3}{#4}
\tikHc{#1/2}{#2-1}{#3+#1}{#4+#1}}
}

\newcommand{\tikHb}[4]{
\draw (0+#3,#1+#4) rectangle +(#1,#1);
\draw (#1+#3,0+#4) rectangle +(#1,#1);
\pgfmathtruncatemacro{\truncr}{#2}
\ifthenelse{\truncr=0}{}{\tikHb{#1/2}{#2-1}{#3+#1}{#4+#1}}
}

\newcommand{\tikHc}[4]{
\draw (0+#3,#1+#4) rectangle +(#1,#1);
\draw (#1+#3,0+#4) rectangle +(#1,#1);
\pgfmathtruncatemacro{\truncr}{#2}
\ifthenelse{\truncr=0}{}{\tikHc{#1/2}{#2-1}{#3}{#4}}
}

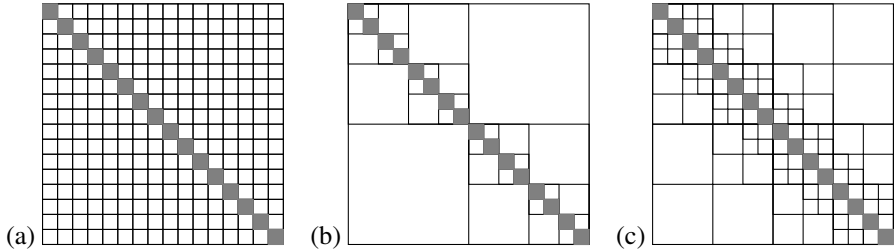
\begin{figure}[h!]
\centerline{
(a)
\begin{tikzpicture}[baseline=0, scale=0.08] 
\draw(0,0) rectangle +(40,40);
\foreach \i in {0,...,15}{
\foreach \j in {0,...,15}{
\draw(2.5*\i,40-2.5*\j) rectangle +(2.5,-2.5);}}
\foreach \i in {0,...,15}{
\filldraw[gray] (2.5*\i,40-2.5*\i) rectangle +(2.5,-2.5);}
\end{tikzpicture}
~~
%
(b)
\begin{tikzpicture}[baseline=0, scale=0.08]
\draw(0,0) rectangle +(40,40);
\tikquad{20}{3}{0}{0}
\end{tikzpicture}
~~
%
(c)
\begin{tikzpicture}[baseline=0, scale=0.08]
\draw(0,0) rectangle +(40,40);
\tikHa{20}{3}{0}{0}
\end{tikzpicture}
%
}
\caption{\label{fig: hierarchical} 
(a) In a simple blocked low rank approximation the diagonal blocks are dense (gray), whereas the 
off-diagonal blocks are low rank. 
(b) In an HODLR matrix the low rank off-diagonal blocks form a hierarchical structure leading 
to a much more compact representation.  
(c) \m{\mathcal{H}^2} matrices are a refinement of this idea. 
}
\end{figure} 
\section{Proofs}

\begin{pfof}{Proposition 1}
Assume that \m{K_{\ell-1}} is spsd. Stage \m{\ell} transforms \m{K_{\ell-1}} to \m{H_\ell=K_\ell\oplus D_\ell}. 
Here \m{K_\ell} is a submatrix of \m{\wbar{H_\ell}=\wbar{Q_\ell}\,\wbar{K_{\ell-1}}\,\wbar{Q_\ell}^\top}, therefore 
it is spsd. \m{D_\ell} is a diagonal matrix that is just the concatenation of the diagonal parts of the 
local compressions, therefore it is also spsd. By induction, if \m{K_0=K}, then all \m{K_\ell} and 
\m{D_\ell} matrices are spsd, and therefore the entire factorization (8) is spsd.
\end{pfof}

\begin{pfof}{Proposition 2}
The cost of computing the compressions in a given stage is at most
\m{b_{\text{max}} sc_{\text{comp}} m_{\text{max}}^{\alpha_{\text{comp}}}}. 
However, since \m{b_{\text{max}} m_{\text{max}}\leq n}, this is upper bounded by 
\m{sc_{\text{comp}} m_{\text{max}}^{\alpha_{\text{comp}}-1}n}. 
The other component is the number of operations required to perform the rotations in each stage. 
The matrix \m{K_{\ell-1}} has to be rotated from both the right and the left by  
\m{\wbar{Q_\ell}=\bigoplus Q^\ell_i}, but since each row of these matrices is \m{c_{sp}} sparse, the total 
per-stage complexity is bounded by \m{2c_{sp}n}. 
\end{pfof}

\begin{pfof}{Proposition 3}
The final core-diagonal matrix $H_s$ has a core of size \m{d_\text{core}}. 
Along with the remaining terms along the diagonal, this factor requires \m{d_\text{core}^2 +n-d_{\text{core}}} storage. 
For each of the \m{s} levels of the MKA, each factor \m{Q_\ell} has \m{n} rows, 
each of which is \m{c_\mathrm{sp}}-sparse, so each of the \m{s} factors has at most 
\m{c_\mathrm{sp} n} nonzero entries. 
Adding everything up yields an upper bound of \m{(sc_\mathrm{sp}+1)n + d_\mathrm{core}^2}.
\end{pfof}


\begin{pfof}{Proposition 5}
As explained above each \m{Q^\ell_i} from an MMF-compression is a product of at most 
\m{\lfloor \gamma m \rfloor} Givens rotations, requiring \m{2\lfloor \gamma m \rfloor} storage. 
All the \m{Q^\ell_i} matrices in a given stage add up to \m{\sum_i 2\lfloor \gamma m^\ell_i \rfloor\leq 2 n} storage.
The size of \m{H_s} is the same as in Proposition 3. 
\end{pfof}

\begin{pfof}{Proposition 6 (sketch)}
\m{\tilde K\V z} is computed by multplying \m{\V z} by each of the factors in (8), from right to left. 
Since each \m{Q^\ell_i} is the product of at most \m{m} Givens rotations, multiplying the corresponding 
block of a vector by \m{Q^\ell_i} has complexity \m{2m}. 
Thus the complexity of multiplying a vector by \m{\wbar{Q_\ell}=\bigoplus Q^\ell_i} is at most \m{2n}. 
There are \m{s} stages on the right of \m{H_s} \m and \m{s} stages on the left, leading to a bound of \m{2sn}. 
Multiplying a vector by \m{H} itself has complexity at most \m{d_{\text{core}}^2+n}.
\end{pfof}

\begin{pfof}{Proposition 7}
All of the matrix operations described in this procedure can boil down to computing a complete eigenvector decomposition (EVD) of \m{\tilde K} and performing matrix operations on the resulting eigenvalues of the decomposition.
\begin{enumerate}
\item \m{\tilde K^\alpha = \sum_{i=1}^n \lambda_i^\alpha v_iv_i^T} where \m{ \{v_i\}_{i=1}^n} is an orthonormal basis of \m{\tilde K}. Since \m{\tilde K = Q_1^T Q_2^T \cdots Q_s^T HQ_s \cdots Q_2 Q_1}, it suffices to compute an EVD of \m{H} which is \m{d_\mathrm{core}}-core-diagonal. To compute the EVD it suffices to compute the EVD of \m{[H]_{[d_\mathrm{core}],[d_\mathrm{core}]}}, which requires $d_\mathrm{core}^3$ operations. Once the EVD is computed, to take the power of the eigenvalues requires only $n$ operations. All together, this is $O(n+d_\mathrm{core}^3)$ operations.
\item \m{ \exp(\beta \tilde K) = \sum_{i=1}^n \exp(\beta \lambda_i) v_iv_i^T} with notation as in \m{\tilde K^\alpha}. Again, the calculation of the EVD of \m{\tilde K} costs $d_\mathrm{core}^3$ operations and the additional procedure of exponentiating $\beta$ times the eigenvalues  takes $2n$ operations. Together this costs $O(n+d_\mathrm{core}^3)$ operations.
\item \m{\mathrm{det}(\tilde K) = \prod_{i=1}^n \lambda_i}. Every rotation matrix has a determinant equal to one, so the $Q_l$ terms which are block-rotation matrices, will also have determinant equal to one. Computing the determinant of \m{H} again boils down to computing the EVD and then taking the product of the eigenvalues. This will also have $O(n+d_\mathrm{core}^3)$.
\end{enumerate}
\end{pfof}
\section{Algorithm}

The pseudocode of the proposed \textbf{Multiresolution Kernel Approximation (MKA)} algorithm is shown in Algorithm~\ref{alg: mka}. MKA is a meta-algorithm, in the sense that it can be used in conjunction with different core-diagonal compressors. 

\begin{algorithm*}[t]
\begin{algorithmic}
\STATE \textbf{Input:} an spsd kernel matrix \m{K\tin\RR^{n\times n}}
\STATE \m{K_0\<\leftarrow K}
\STATE \textbf{for} (\m{\ell\<=1} to \m{s})\,\m{\{}
\STATE ~~~ \textbf{cluster}~ the columns of \m{K_{\ell-1}} into \m{(\Ccal_{1}^\ell, \dots, \Ccal_{p_\ell}^\ell)}
\STATE ~~~ \textbf{permute}~ the rows/columns of \m{K_{\ell-1}} 
according to \m{(\Ccal_1^{\ell}, \dots, \Ccal_{p_{\ell}}^{\ell})} to get \m{\wbar{K_{\ell-1}}}
\STATE ~~~ \textbf{for} (\m{i \<= 1} to \m{p_\ell} )\, \m{\{ }
\STATE ~~~ ~~~\m{(Q_i^\ell,c^\ell_i) \leftarrow} \textsc{Compress}(\m{[\wbar{K}_{\ell-1}]_{i,i}})
\STATE ~~~ \m{\}}
\STATE ~~~ \m{\wbar{Q}_\ell\leftarrow \bigoplus_i Q_i^\ell}\\
\STATE ~~~ \m{\wbar{H}_\ell \leftarrow \wbar{Q}_\ell \wbar K_{\ell-1}  \wbar{Q}_\ell^\top}
\STATE ~~~ \m{c_\ell=\sum_{i=1}^{p_i} c^\ell_i}
\STATE ~~~ \textbf{permute}~ the rows/columns of \m{\wbar{H}_\ell} so that the cores appear in the top left \m{c_\ell\times c_\ell} submatrix to get \m{H_\ell}
\STATE ~~~ \m{K_\ell\leftarrow [H_\ell]_{1:c_\ell,1:c_\ell}} \hfill // this is the ``core'' part of \m{H_\ell}
\STATE ~~~\m{D_\ell\leftarrow \diag(\diag([H_\ell]_{c_\ell+1:, c_\ell+1,:}))} \hfill // this is the ``diagonal'' part of \m{H_\ell}
\STATE \m{\}}
\STATE \textbf{Output:} \m{(\wbar{Q_1},\ldots,\wbar{Q_s},\seq{D}{s},K_s)}
\end{algorithmic}
\caption{The MKA algorithm. \textsc{Compress} is any suitable core/diagonal compression 
routine, e.g., a Jacobi MMF.}\label{alg: mka} 
\end{algorithm*}
\begin{table}[]
\centering
\caption{\label{tbl: datasets}Summary of the datasets used in our experiments}
\begin{tabular}{l l l l}
\toprule 
Dataset      & Size     & Dimensions 
\\
\hline
housing      & 506      & 13         
\\
rupture      & 2066     & 30         
\\
wine         & 4898     & 11         
\\
pageblocks   & 5473     & 10         
\\
compAct      & 8192     & 21         
\\
pendigit     & 10992    & 16         
\\
\bottomrule
\end{tabular}
\vspace{-10pt}
\end{table}

\section{Experiments}

\subsection{Datasets}
We used six data sets in our experiments, of which, \texttt{rupture} is from Materials algorithms project program \footnote{https://www.phase-trans.msm.cam.ac.uk/map/map.html} and the others are from UCI machine learning repository \footnote{https://archive.ics.uci.edu/ml/datasets.html}. The detailed summary of the datasets is in Table~\ref{tbl: datasets}. 

\begin{figure}[t!]
\centering
\begin{minipage}[b]{0.24\textwidth}
	\centering  \tiny{\textbf{wine}}\\
		\includegraphics[width=1\textwidth]{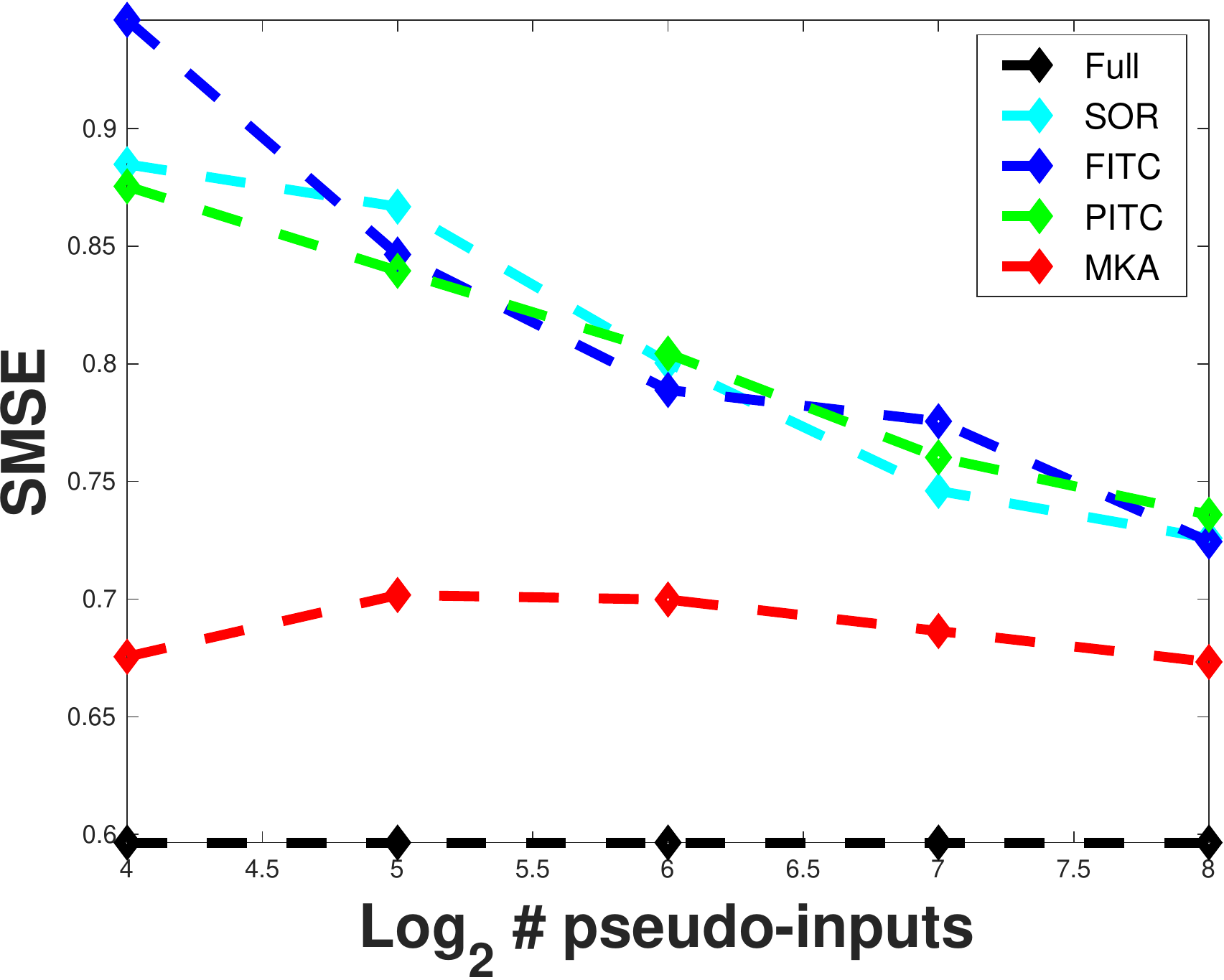}
\end{minipage}
\begin{minipage}[b]{0.24\linewidth}
	\centering \tiny{\textbf{wine}}\\
\includegraphics[width=1\textwidth]{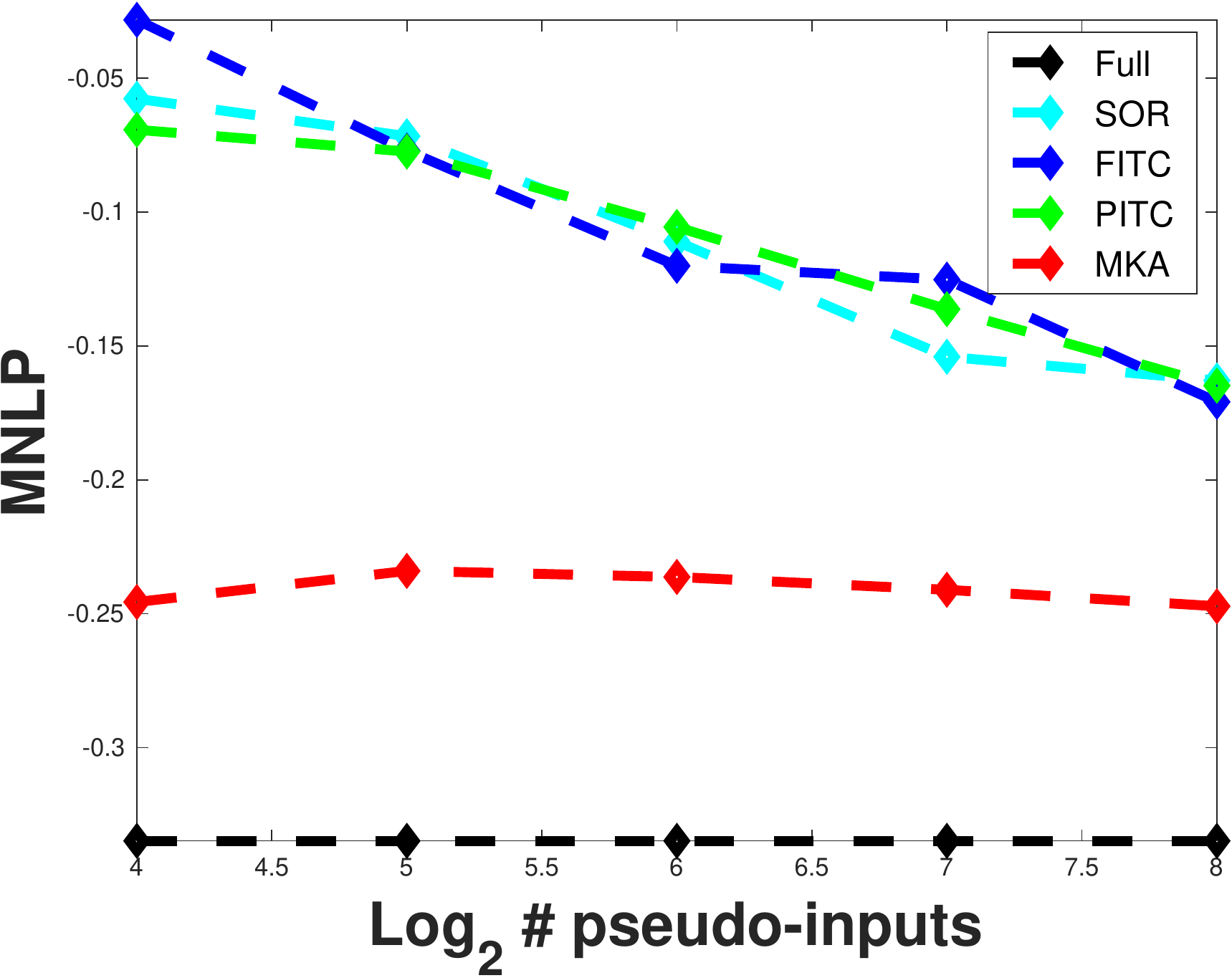}
\end{minipage} 
\begin{minipage}[b]{0.24\linewidth}
	\centering \tiny{\textbf{pageblocks}}\\
\includegraphics[width=1\textwidth]{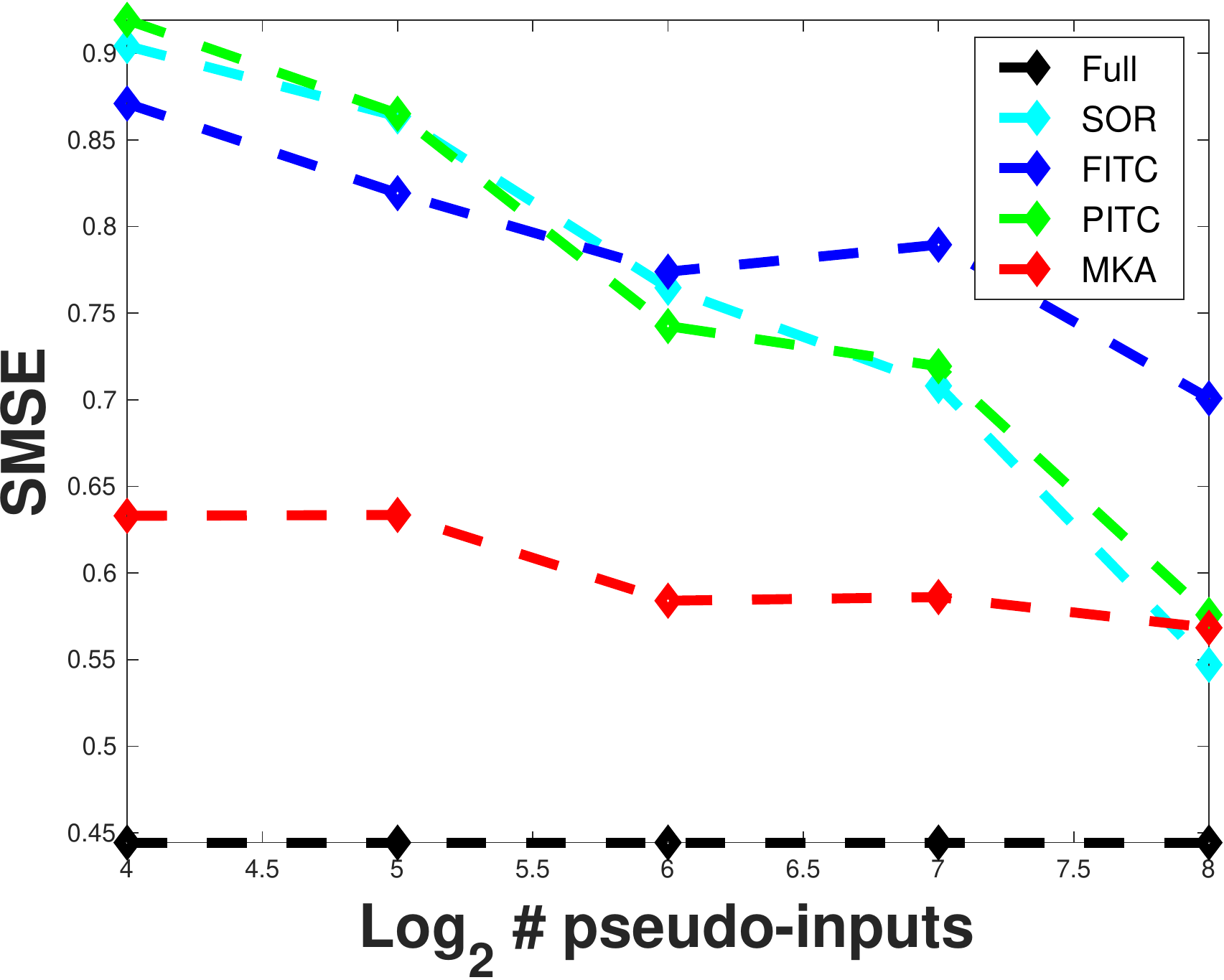}
\end{minipage}
\begin{minipage}[b]{0.24\linewidth}
	\centering \tiny{\textbf{pageblocks}}\\
\includegraphics[width=1\textwidth]{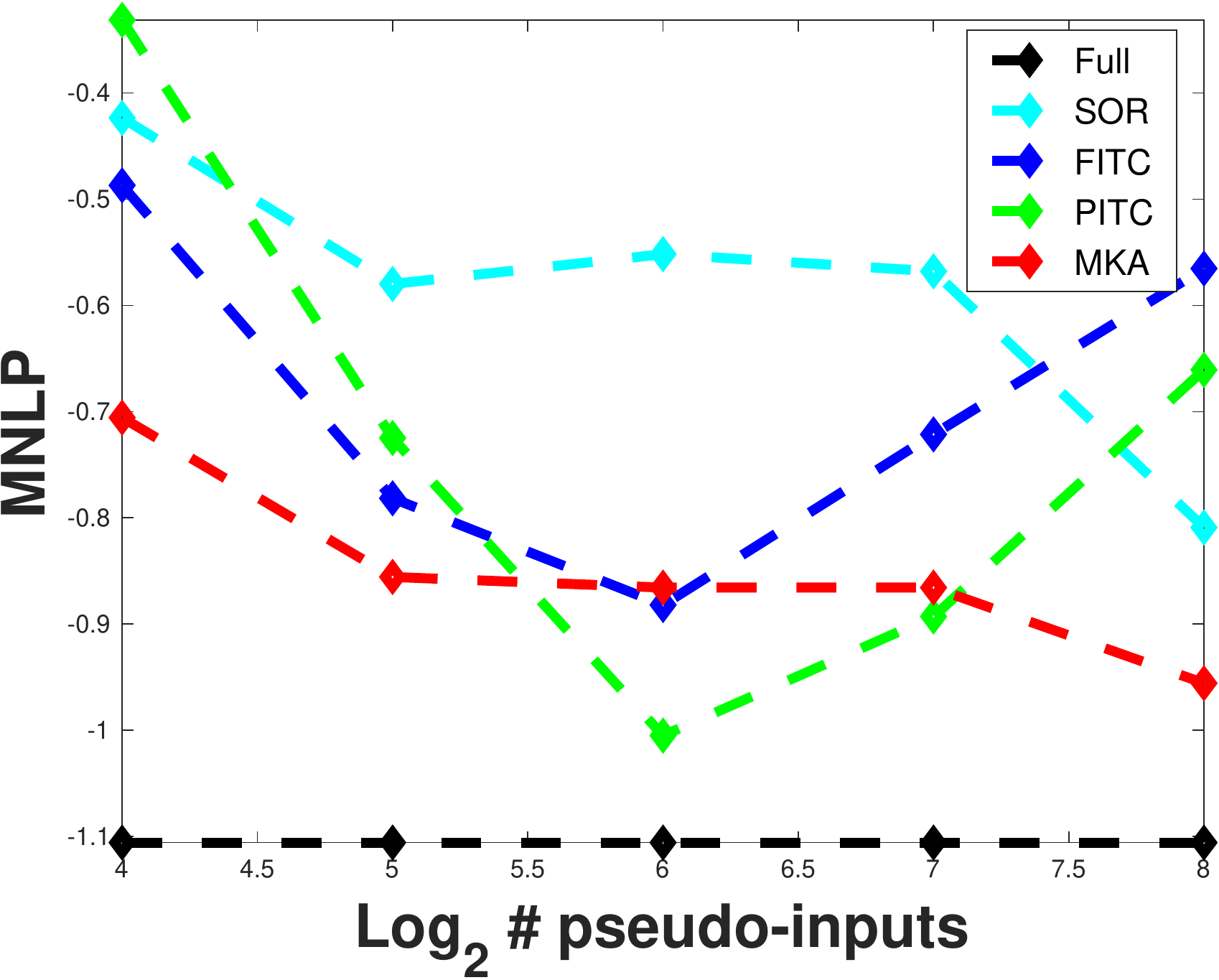}
\end{minipage} 
\begin{minipage}[b]{0.24\linewidth}
	\centering  \tiny{\textbf{compAct}}\\
\includegraphics[width=1\textwidth]{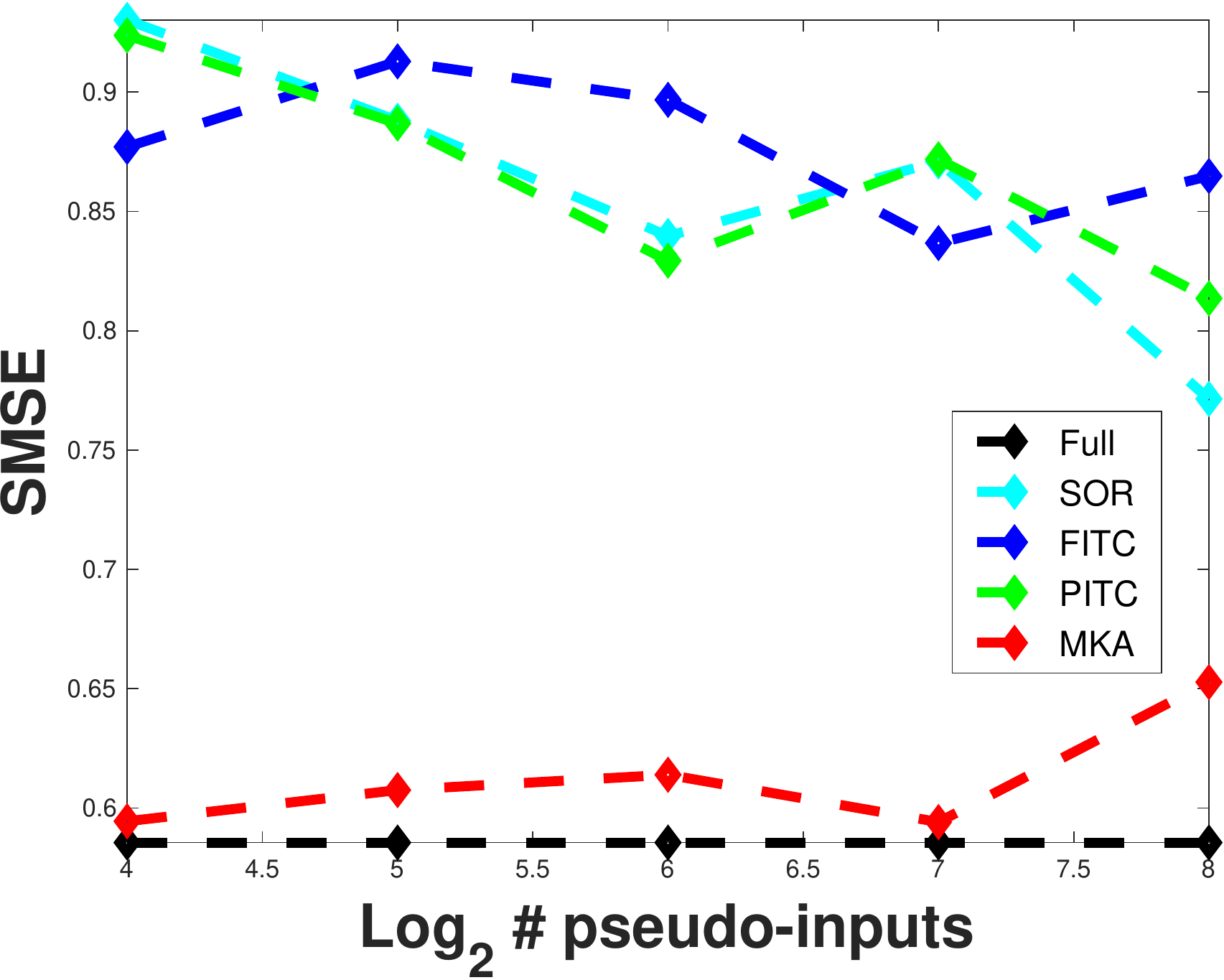}
\end{minipage}
\begin{minipage}[b]{0.24\linewidth}
	\centering \tiny{\textbf{compAct}}\\
\includegraphics[width=1\textwidth]{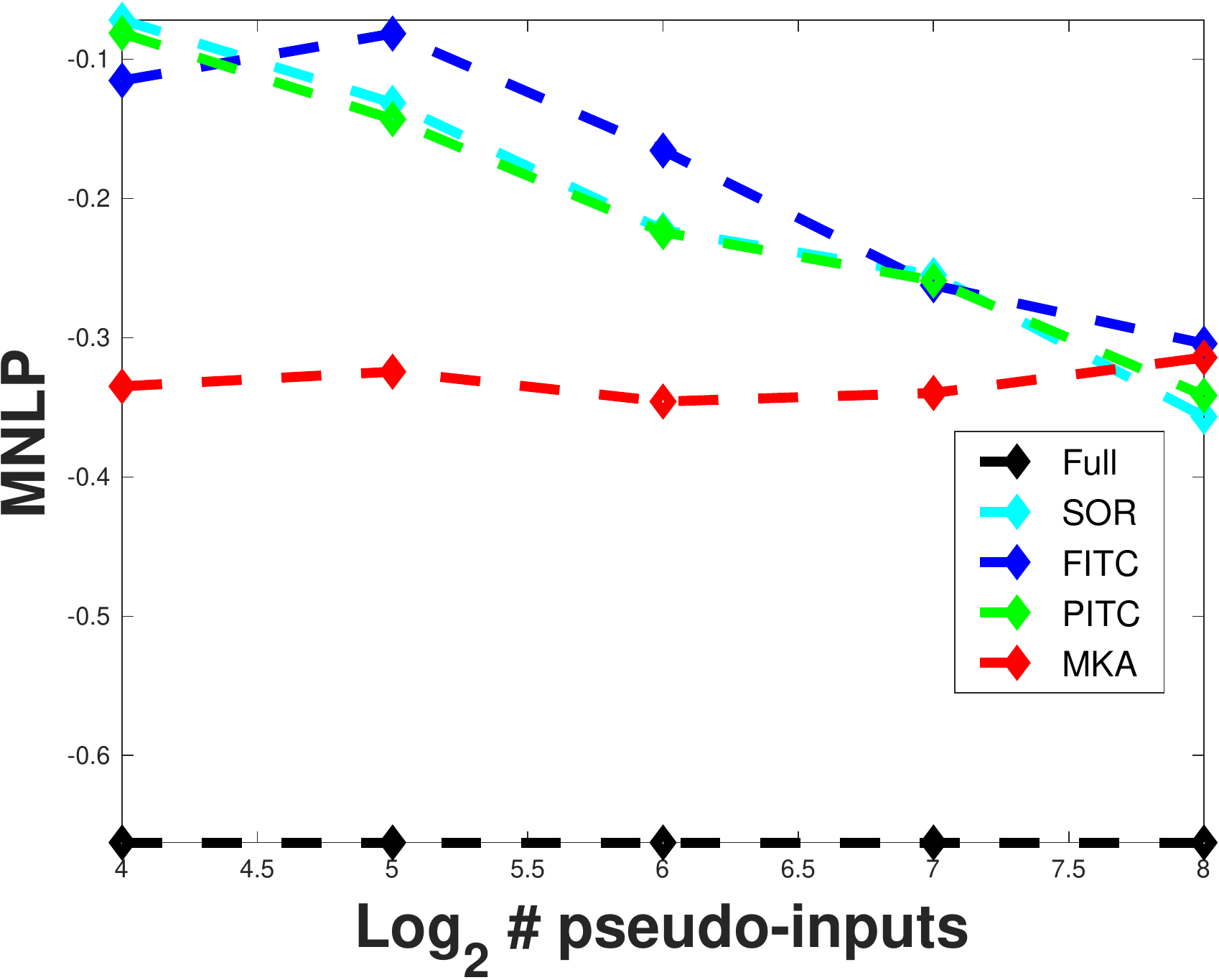}
\end{minipage} 
\begin{minipage}[b]{0.24\linewidth}
	\centering \tiny{\textbf{pendigit}}\\
\includegraphics[width=1\textwidth]{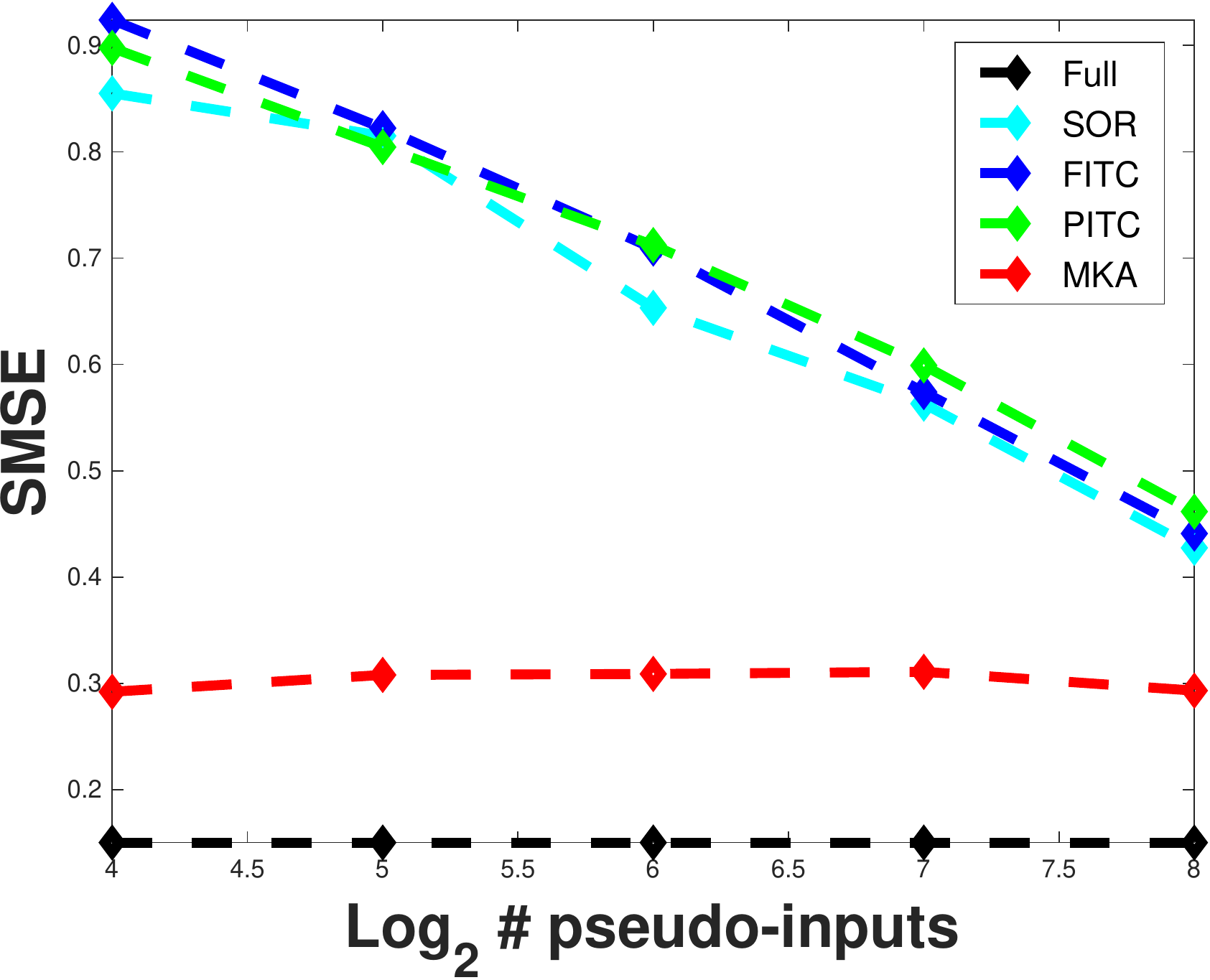}
\end{minipage}
\begin{minipage}[b]{0.24\linewidth}
	\centering \tiny{\textbf{pendigit}}\\
\includegraphics[width=1\textwidth]{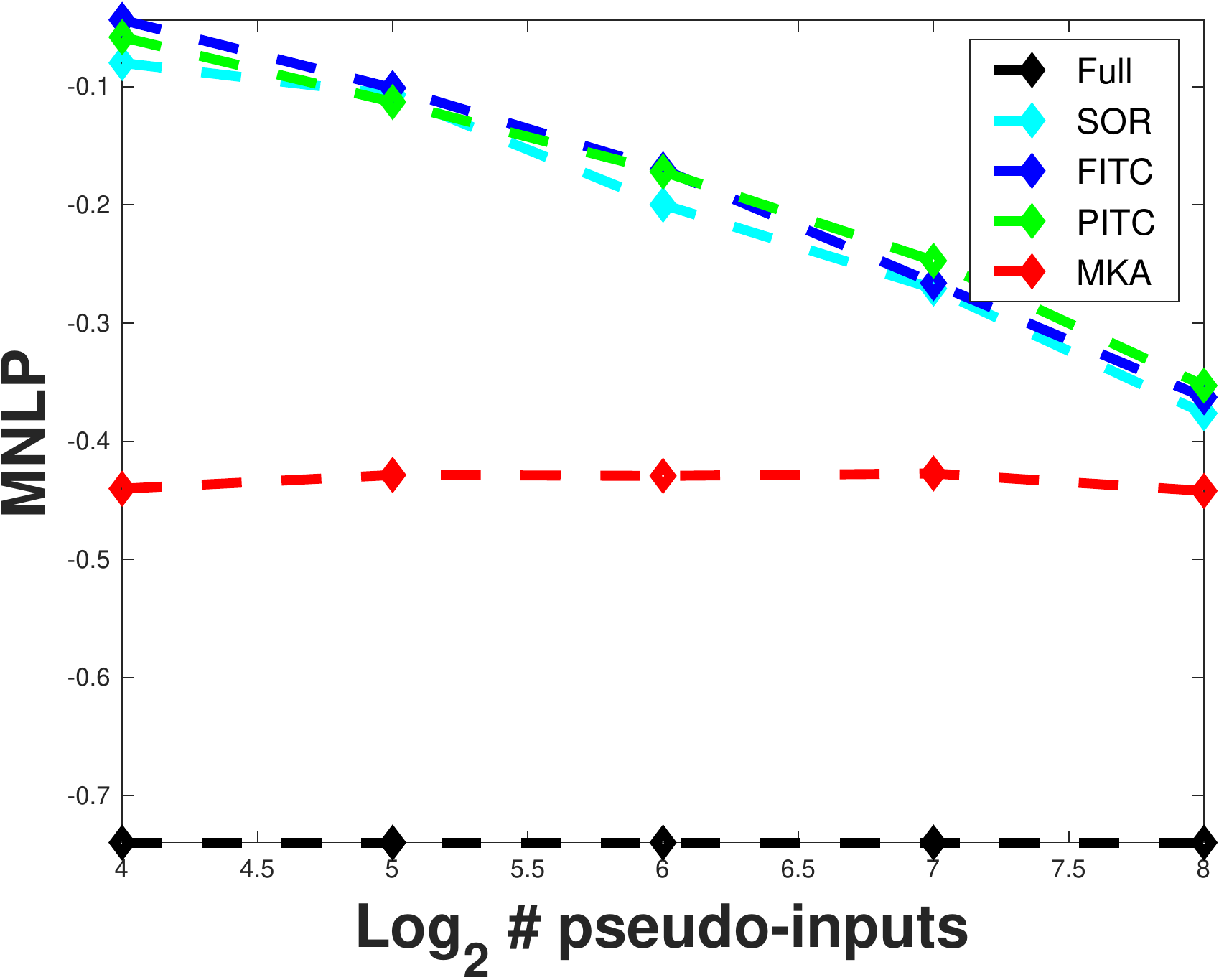}
\end{minipage}
\vspace{-6pt}\mbox{} \\
\caption{SMSE and MNLP as a function of the number of pseudo-inputs/\m{d_{\text{core}}} on the rest four datasets. 
	In the given range MKA clearly outperforms the other methods in both error measures.
}\label{fig:rank-supp}  \vspace{-0.1in}
\end{figure}

\subsection{Evaluation as a function of number of pseudo-inputs/\m{d_{\text{core}}}}
We compare regression results in terms of both the predictive mean and variance (i.e. SMSE/MNLP) as a function of the number of pseudo-inputs/ \m{d_{\text{core}}}, which represents the approximation/compression level of the kernel matrix. 
Across the range of pseudo-inputs/\m{d_{\text{core}}} considered in Figure~\ref{fig:rank-supp} for selected data sets in this supplementary material , MKA outperformed other methods in terms of both prediction accuracy (SMSE) and variance assessment (MNLP), whereas for other methods more error was accumulated as fewer pseudo-inputs were used. These results on additional data sets to those illustrated in the main paper confirm the position that MKA is in many cases a superior method for kernel matrix compression. In these four data sets, as well, MKA's performance was nearly constant across different sizes of \m{d_{\text{core}}} -- likely due to the information preserved along the main diagonal in the $c$-core diagonal matrix of MKA’s kernel matrix approximation. Moreover, the results for MEKA on the selected data sets are absent due to the fact that the approximate kernel matrix found by MEKA for these data sets loses the spsd property, and thus fails to show prediction results in the experiments.